\documentclass[twoside,11pt]{article}

\usepackage{jmlr2e}
\usepackage{bm}
\usepackage{verbatim}
\usepackage{amsmath}
\usepackage{array, multirow}
\usepackage{graphicx}

\usepackage{makecell}%To keep spacing of text in tables
\setcellgapes{4pt}%parameter for the spacing

\usepackage{xcolor}
\usepackage[normalem]{ulem}

\usepackage{lastpage}
\jmlrheading{21}{2020}{1-\pageref{LastPage}}{3/20; Revised
7/20}{7/20}{20-212}{Sanmit Narvekar, Bei Peng, Matteo Leonetti, Jivko Sinapov, Matthew E.~Taylor, and Peter Stone}

\ShortHeadings{Curriculum Learning for Reinforcement Learning Domains}{Narvekar, Peng, Leonetti, Sinapov, Taylor, and Stone}
\firstpageno{1}

\begin{document}

\title{Curriculum Learning for Reinforcement Learning Domains: A Framework and Survey}

\author{\name Sanmit Narvekar \email sanmit@cs.utexas.edu \\ 
\addr Department of Computer Science \\ University of Texas at Austin
\AND
\name Bei Peng \email bei.peng@cs.ox.ac.uk \\
\addr Department of Computer Science \\ University of Oxford
\AND
\name Matteo Leonetti \email m.leonetti@leeds.ac.uk \\
\addr School of Computing \\ University of Leeds
\AND
\name Jivko Sinapov \email jivko.sinapov@tufts.edu \\
\addr Department of Computer Science \\ Tufts University
\AND
\name Matthew E.~Taylor \email matthew.e.taylor@ualberta.ca \\
\addr Alberta Machine Intelligence Institute\\
Department of Computing Science\\
University of Alberta
\AND
\name Peter Stone \email pstone@cs.utexas.edu \\
\addr Department of Computer Science \\ University of Texas at Austin \\
and Sony AI
}

\editor{George Konidaris}

\maketitle

\begin{abstract}%
Reinforcement learning (RL) is a popular paradigm for addressing sequential decision tasks in which the agent has only limited environmental feedback. Despite many advances over the past three decades, learning in many domains still requires a large amount of interaction with the environment, which can be prohibitively expensive in realistic scenarios. To address this problem, transfer learning has been applied to reinforcement learning such that experience gained in one task can be leveraged when starting to learn the next, harder task. More recently, several lines of research have explored how tasks, or data samples themselves, can be sequenced into a \emph{curriculum} for the purpose of learning a problem that may otherwise be too difficult to learn from scratch. In this article, we present a framework for curriculum learning (CL) in reinforcement learning, and use it to survey and classify existing CL methods in terms of their assumptions, capabilities, and goals. Finally, we use our framework to find open problems and suggest directions for future RL curriculum learning research.  

\end{abstract}


\section{Introduction}
\label{sec:Intro}
%%%%%%%%%%%%%%%%

Curricula are ubiquitous throughout early human development, formal education, and life-long learning all the way to adulthood. Whether learning to play a sport, or learning to become an expert in mathematics, the training process is organized and structured so as to present new concepts and tasks in a sequence that leverages what has previously been learned. In a variety of human learning domains, the quality of the curricula has been shown to be crucial in achieving success. Curricula are also present in animal training, where it is commonly referred to as shaping \citep{skinner1958reinforcement,peterson2004day}. 

\begin{figure}
\centering
  \includegraphics[width=\linewidth]{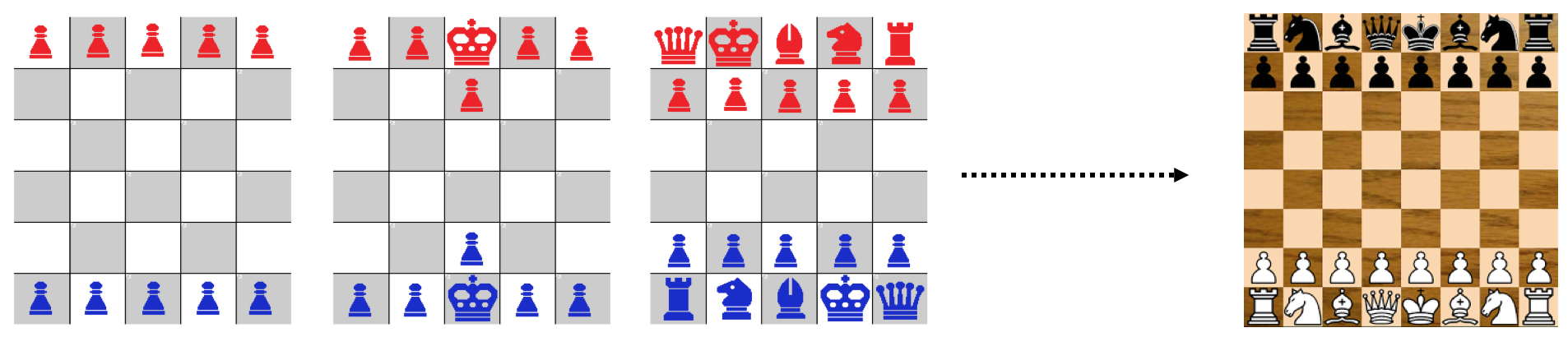}
\caption{Different subgames in the game of Quick Chess, which are used to form a curriculum for learning the full game of Chess.}
  \label{fig:quickchess}
\end{figure}

As a motivating example, consider the  game  of  Quick  Chess (shown in Figure \ref{fig:quickchess}), a  game  designed  to  introduce  children  to  the  full  game of chess, by using a sequence of progressively more difficult ``subgames.'' For example, the first subgame is played on a 5x5 board with only pawns, where the player learns how pawns move, get promoted, and take other pieces. Next, in the second subgame, the king piece is added, which introduces a new objective: keeping the king alive.  In each successive subgame, new elements are introduced (such as new pieces, a larger board, or different configurations) that require learning new skills and building  upon  knowledge  learned  in  previous  games.   The final game is the full game of chess.

The idea of using such curricula to train artificial agents dates back to the early 1990s, where the first known applications were to grammar learning \citep{elman1993learning,rohde1999language}, robotics control problems \citep{sanger1994neural}, and classification problems \citep{bengio2009curriculum}. Results showed that the order of training examples matters and that generally, incremental learning algorithms can benefit when training examples are ordered in increasing difficulty. The main conclusion from these and subsequent works in curriculum learning is that starting small and simple and gradually increasing the difficulty of the task can lead to faster convergence as well as increased performance on a task. 

Recently, research in reinforcement learning (RL) \citep{Sutton98} has been exploring how agents can leverage transfer learning \citep{Lazaric08,Taylor09} to re-use knowledge learned from a source task when attempting to learn a subsequent target task. As knowledge is transferred from one task to the next, the sequence of tasks induces a curriculum, which has been shown to improve performance on a difficult problem and/or reduce the time it takes to converge to an optimal policy. 

Many groups have been studying how such a curriculum can be generated automatically to train reinforcement learning agents, and many approaches to do so now exist. However, what exactly constitutes a curriculum and what precisely qualifies an approach as being an example of curriculum learning is not clearly and consistently defined in the literature. There are many ways of defining a curriculum: for example, the most common way is as an ordering of tasks. At a more fundamental level, a curriculum can also be defined as an ordering of individual experience samples. In addition, a curriculum does not necessarily have to be a simple linear sequence. One task can build upon knowledge gained from multiple source tasks, just as courses in human education can build off of multiple prerequisites. 

Methods for curriculum generation have separately been introduced for areas such as robotics, multi-agent systems, human-computer and human-robot interaction, and intrinsically motivated learning. This body of work, however, is largely disconnected. In addition, many landmark results in reinforcement learning, from TD-Gammon \citep{tesauro1995temporal} to AlphaGo \citep{silver2016mastering} have implicitly used curricula to guide training. In some domains, researchers have successfully used methodologies that align with our definition of curriculum learning without explicitly describing it that way (e.g., self-play).
%curricula can be obtained for ``free," (e.g., self-play) often without researchers realizing that it is a curriculum. 
Given the many landmark results that have utilized ideas from curriculum learning, we think it is very likely that future landmark results will also rely on curricula, perhaps more so than researchers currently expect.
Thus, having a common basis for discussion of ideas in this area is likely to be useful for future AI challenges. 

\subsection*{Overview}

The goal of this article is to provide a systematic overview of curriculum learning (CL) in RL settings and to provide an over-arching framework to formalize this class of methods. We aim to define classification criteria for  computational models of curriculum learning for RL agents, that describe the curriculum learning research landscape over a broad range of frameworks and settings. The questions we address in this survey include:
%that distinguish curriculum learning methods include:

\begin{itemize}

\item What is a \emph{curriculum}, and how can it be represented for reinforcement learning tasks? At the most basic level, a curriculum can be thought of as an ordering over experience samples. However, it can also be represented at the task level, where a set of tasks can be organized into a sequence or a directed acyclic graph that specifies the order in which they should be learned. We address this question in detail in Section \ref{sec:curricula_def}.

\item What is the \emph{curriculum learning} method, and how can such methods be evaluated? We formalize this class of methods in Section \ref{sec:cl_def} as consisting of three parts, and extend metrics commonly used in transfer learning (introduced in Section \ref{sec:background}) to the curriculum setting to facilitate evaluation in Section \ref{sec:cl_eval}.  

%We formalize all of these in Section \ref{sec:curriculum_problem} in a way that encompasses a broad variety of curriculum based approaches. 

%\item What are the goals of the curriculum learning method, and by what metric(s) will success be measured? Section \ref{sec:background} examines commonly used metrics, often adopted from transfer learning, as well as different settings where the use of curricula can improve learning.

%\item How are curricula represented and what are the different structural forms that they may take? For example, a sequence of tasks can be seen as a curriculum but tasks may also be arranged into a directed acyclic graph, or a tree, as discussed in Section \ref{sec:curriculum_problem}.

\item How can tasks be constructed for use in a curriculum? The quality of a curriculum is dependent on the quality of tasks available to select from. Tasks can either be generated in advance, or dynamically and on-the-fly with the curriculum. Section \ref{sec:task_generation} surveys works that examine how to automatically generate good intermediate tasks. 

%the most common assumptions and means by which tasks are generated (e.g., reducing the size of the state-space or action-space).

\item How can tasks or experience samples be sequenced into a curriculum? In practice, most curricula for RL agents have been manually generated for each problem. However, in recent years, automated methods for generating curricula have been proposed. Each makes different assumptions about the tasks and transfer methodology used. In Section \ref{sec:task_sequencing}, we survey these different automated approaches, as well as describe how humans have approached curriculum generation for RL agents.

%while most uses of curricula for RL agents involve manual curriculum construction for a particular problem, some principled approaches for automatically constructing a curriculum have also been proposed. Section \ref{sec:task_sequencing} enumerates the possibilities ranging from fully manual to fully automatic curriculum design, while Section \ref{sec:curriculum_with_human} describes how naive users can also construct curricula for RL agents. 

\item How can an agent transfer knowledge between tasks as it learns through a curriculum?
%What are the types of knowledge transfer that an agent employs as it learns tasks through a curriculum? 
Curriculum learning approaches make use of transfer learning methods when moving from one task to another. Since the tasks in the curriculum can vary in state/action space, transition function, or reward function, it's important to transfer relevant and reusable information from each task, and effectively combine information from multiple tasks. Methods to do this are enumerated and discussed in Section \ref{sec:knowledge_transfer}.

\end{itemize}

The next section provides background in reinforcement learning and transfer learning. In Section \ref{sec:curriculum_problem}, we define the curriculum learning method, evaluation metrics, and the dimensions along which we will classify curriculum learning approaches. Section \ref{sec:CLforRLAgents}, which comprises the core of the survey, provides a detailed overview of the existing state of the art in curriculum learning in RL, with each subsection considering a different component of the overall curriculum learning approach. %and means by which existing methods can be distinguished and classified. %It also looks closely at existing methods for automated curriculum construction for human students and relates those methods to possible counterparts in the RL literature. 
Section \ref{sec:related_work} discusses paradigms related to curriculum learning for RL, such as curriculum learning for supervised learning and for human education. Finally, in Section \ref{sec:open_problems}, we identify gaps in the existing literature, outline the limitations of existing CL methods and frameworks, and provide a list of open problems.

\section{Background}
\label{sec:background}
%%%%%%%%%%%%%%%%%

In this section, we provide background on Reinforcement Learning (RL) and Transfer Learning (TL). %, and formalize the problem of Curriculum Learning in RL. 

\subsection{Reinforcement Learning}

Reinforcement learning considers the problem of how an agent should act in its environment over time, so as to maximize some scalar reward signal. We can formalize the interaction of an agent with its environment (also called a \emph{task}) as a Markov Decision Process (MDP). In this article, we restrict our attention to \emph{episodic} MDPs:\footnote{In continuing tasks, a discount factor $\gamma$ is often included. For simplicity, and due to the fact that tasks typically terminate in curriculum learning settings, we present the undiscounted case. But unless otherwise noted, our definitions and
discussions can easily apply to the discounted case as well.}

%MDPs are used at two different levels in this paper - one for modeling individual tasks, and one for modeling the task sequencing process. We introduce the formulation for individual tasks here. %, and describe the task selection process MDP in Section \ref{sec:cmdp}. 

%an agent with environment, and one with learning process. In this section, we introduce the standard use of MDPs with environment. Sec 3 formulates learning process as higher MDP. 
%  Describe tasks as MDP and introduce MDP
%In the standard formulation, 

\begin{definition}

An episodic MDP $M$ is a 6-tuple $(\mathcal{S}, \mathcal{A}, p, r, \Delta s_0, \mathcal{S}_f)$, where $\mathcal{S}$ is the set of states,
%in the environment
 $\mathcal{A}$ is the set of actions,
%the agent can take
 $p(s' | s, a)$ is a transition function that gives the probability of transitioning to state $s'$ after taking action $a$ in state $s$, and $r(s, a, s')$ is a reward function that gives the immediate reward for taking action $a$ in state $s$ and transitioning to state $s'$. In addition, we shall use $\Delta s_0$ to denote the initial state distribution, and $\mathcal{S}_f$ to denote the set of terminal states. 
%In addition, we shall use $S_0$ to denote the set of initial states, and $S_f$ to denote the set of terminal states. 

\end{definition}

%\comments{I still like the idea of using $p_0$ over $\Delta s_0$ for the initial state distribution. But I don't think we ever use this notation later.}

We consider time in discrete time steps. At each time step $t$, the agent observes its state and chooses an action according to its \emph{policy} $\pi(a|s)$. The goal of the agent is to learn an \emph{optimal policy} $\pi^*$, which maximizes the  expected %long term sum of rewards 
\emph{return} $G_t$ (the cumulative sum of rewards $R$) until the episode ends at timestep $T$: 
\begin{equation*}
G_t = \sum_{i=1}^{T-t} R_{t+i}
\end{equation*}

% \commentm{I think we can make this shorter. This is an introduction to RL at a very basic level, we can probably assume RL knowledge from the reader, we mostly need to clarify the notation} 

There are three main classes of methods to learn $\pi^*$: value function approaches, policy search approaches, and actor-critic methods. In \emph{value function approaches}, a value $v_\pi(s)$ is first learned for each state $s$, representing the expected return achievable from $s$ by following policy $\pi$. Through policy evaluation and policy improvement, this value function is used to derive a policy better than $\pi$, until convergence towards an optimal policy. %For example, if the model (i.e., transition and reward functions) is known, the action that leads to the best combination of reward and next state is selected. 
Using a value function in this process requires a model of the reward and transition functions of the environment. If the model is not known, one option is to learn an action-value function instead, $q_\pi(s,a)$, which gives the expected return for taking action $a$ in state $s$ and following $\pi$ after:

\begin{equation*}
q_\pi(s,a) =  \sum_{s'}p(s'|s,a) [r(s,a,s') + q_\pi(s',a')] \textrm{ , where } a' \sim \pi(\cdot | s')
\end{equation*}

%\comments{I don't really like the idea of saying $a'$ is sampled from $\pi$ either because usually q-learning is with deterministic policies. Would it be ok to just remove the ``where .. " part?}

The action-value function can be iteratively improved towards the optimal action-value function $q_*$ with on-policy methods such as SARSA \citep{Sutton98}. The optimal action-value function can also be learned directly with off-policy methods such as $Q$-learning \citep{Watkins92}. An optimal policy can then be obtained by choosing action $\text{argmax}_a q_*(s,a)$ in each state.
If the state space is large or continuous, the action-value function can instead be estimated using a function approximator (such as a neural network), $q(s,a;\bm{w}) \approx q_*(s,a)$, where $\bm{w}$ are the weights of the network. %as a function of state \emph{features} $\bm{\phi}(s)$ and a weight vector $\bm{\theta}$. %We discuss several common representations for function approximation using such vectors in the next section.  

% is theta always a weight vector of could it be some other model parameters?

%In continuous domains, instead of learning a separate action-value for each state-action pair,   

In contrast, \emph{policy search methods} directly search for or learn a parameterized policy $\pi_{\bm{\theta}}(a|s)$, without using an intermediary value function. Typically, the parameter $\bm{\theta}$ is modified using search or optimization techniques to maximize some performance measure $J(\bm{\theta})$. For example, in the episodic case, $J(\bm{\theta})$ could correspond to the expected value of the policy parameterized by $\bm{\theta}$ from the starting state $s_0 \sim \Delta s_0$: $v_{\pi_\theta}(s_0)$.

A third class of methods, \emph{actor-critic methods}, maintain a parameterized representation of both the current policy and value function. 
The actor is a parameterized policy that dictates how the agent selects actions. The critic estimates the (action-)value function for the actor using a policy evaluation method such as temporal-difference learning. The actor then updates the policy parameter in the direction suggested by the critic. %and compute the optimal policy updates with respect to the estimated values. 
An example of actor-critic methods is Deterministic Policy Gradient~\citep{silver2014deterministic}.

\subsection{Transfer Learning}

%Curriculum learning relies on transferring and generalizing knowledge acquired early on in the training process to later, typically more difficult stages. Typically, this has been considered in the context of pairs of tasks: a \emph{source} and a \emph{target}.

%We denote the task that agent must learn as the \emph{target task}. 
% Standard learning scenario -- i.e. what is no curriculum
In the standard reinforcement learning setting, an agent usually starts with a random policy, and directly attempts to learn an optimal policy for the target task. When the target task is difficult, for example due to adversarial agents, poor state representation, or sparse reward signals, learning can be very slow. 

Transfer learning is one class of methods and area of research that seeks to speed up training of RL agents. %, and lays the foundation for curriculum learning. 
The idea behind transfer learning is that instead of learning on the \emph{target task} tabula rasa, the agent can first train on one or more \emph{source task} MDPs, and \emph{transfer} the knowledge acquired to aid in solving the target. %Various researchers over the past 2 decades have examined how to transfer knowledge in the form of 
%Curriculum learning builds on the assumption that learning one task can speed up learning on another, a concept that's come to be known as \emph{transfer learning}~\cite{Taylor09,Lazaric11}. 
This knowledge can take the form of 
samples \citep{Lazaric08,Lazaric11b}, options \citep{Soni06}, policies \citep{Fernandez10}, models \citep{Fachantidis13}, or value functions \citep{Taylor05}.
As an example, in value function transfer \citep{Taylor07}, the parameters of an action-value function $q_{source}(s, a)$ learned
in a source task are used to initialize the action-value function in the
target task $q_{target}(s, a)$. This biases exploration and action selection in the target task based on experience acquired in the source task. 

% usually the underscore to q refers to a policy

Some of these methods assume that the source and target MDPs either share state and action spaces, or that a \emph{task mapping} \citep{Taylor07} is available to map states and actions in the target task to known states and actions in the source. Such mappings can be specified by hand, or learned automatically \citep{Taylor08,Ammar15}. Other methods assume the transition or reward functions do not change between tasks. The best method to use varies by domain, and depends on the relationship between source and target tasks. Finally, while most methods assume that knowledge is transferred from one source task to one target task, some methods have been proposed to transfer knowledge from several source tasks directly to a single target \citep{svetlik2017automatic}. See \cite{Taylor09} or \cite{lazaric2012transfer} for a survey of transfer learning techniques. 

%\comments{How much detail into different transfer learning methods do we want to go into? Also, for most of my stuff, I've formulated it as episodic MDPs, though we could make that more general. I also think the evaluation metrics stuff should go here, or in a separate section with the formulation material.}

%This assumes that the MDPs either share state and action spaces, or that a task mapping \cite{Taylor07} is available between the MDPs. 

%This assumes that the action-value function must be able to generate a policy for both source and target, or that a mapping is available to do so. 
%Typically this is done by using an egocentric representation for the features. 

\begin{comment}
\begin{figure*}[htbp]
\begin{center}$
\begin{array}{cc}
\includegraphics[height=5cm]{img/mapping.png} & \includegraphics[height=5cm]{img/metrics.png} \\ 
(a) & (b) \\
\end{array}$
\end{center}
\caption{(a) An inter-task mapping from states and actions in the target task to states and actions in a source. (b) Performance metrics for transfer learning.  Images from \citet{Taylor09}.} 
\label{tlImage}
\end{figure*}
\end{comment}

\subsection{Evaluation Metrics for Transfer Learning}
\label{sec:evaluationTL}

%Describe the three basic metrics used in transfer learning: 1) jump start; 2) time to threshold; 3) asymptotic performance.

\begin{figure*}[t]
\begin{center}$
\begin{array}{cc}
\includegraphics[height=4cm]{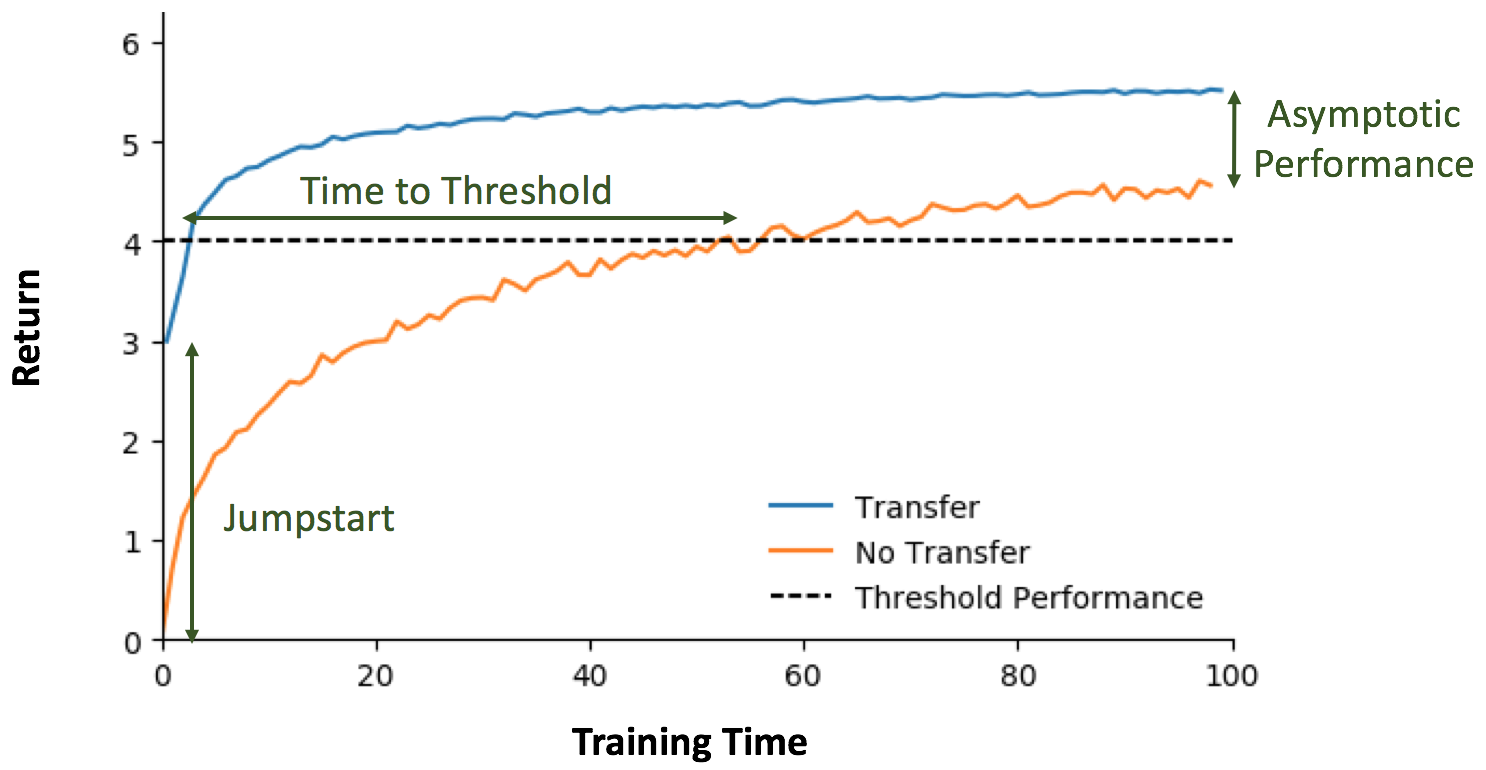} & \includegraphics[height=4cm]{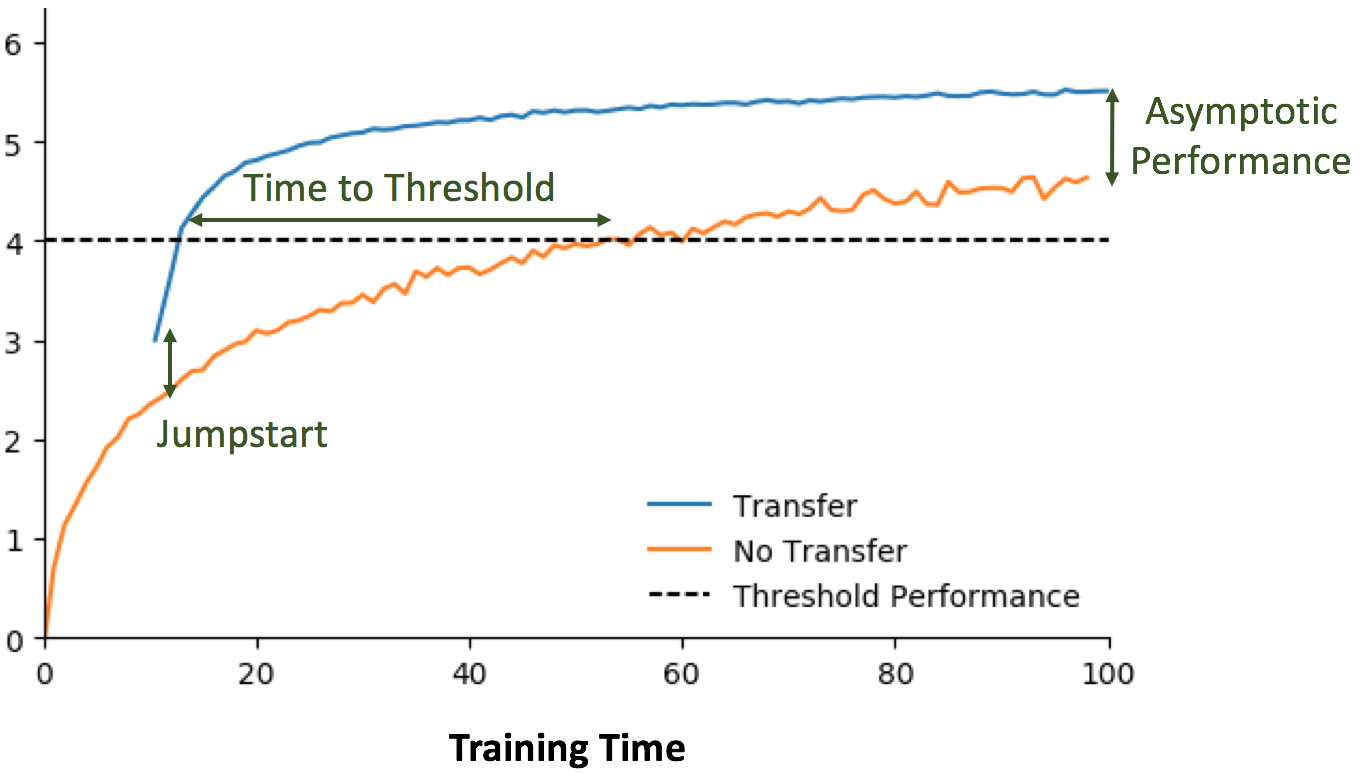} \\
(a) & (b) 
\end{array}$
\end{center}
\caption{Performance metrics for transfer learning using (a) weak transfer and (b) strong transfer with offset curves.}
\label{tlMetrics}
\end{figure*}

There are several metrics to quantify the benefit of transferring from a source task to a target task \citep{Taylor09}.  
Typically, they compare the learning trajectory on the target task for an agent after transfer, with an agent that learns directly on the target task from scratch (see Figure \ref{tlMetrics}a). 
One metric is \emph{time to threshold}, which computes how much faster an agent can learn a policy that achieves expected return $ G_0 \geq \delta$ on the target task if it transfers knowledge, as opposed to learning the target from scratch, where $\delta$ is some desired performance threshold. Time can be measured in terms of CPU time, wall clock time, episodes, or number of actions taken. %; in this work, we will usually measure \emph{time} using the number of actions taken.
Another metric is \emph{asymptotic performance}, which compares the final performance after convergence in the target task of learners when using transfer versus no transfer. The \emph{jumpstart} metric instead measures the initial performance increase on the target task as a result of transfer. 
Finally, the \emph{total reward} ratio compares the total reward accumulated by the agent during training up to a fixed stopping point, using transfer versus not using transfer.

%In order to quantify the benefit of using a curriculum, we extend several metrics devised for transfer learning. 

%In our work, we focus on 2 of these metrics to evaluate the utility of learning via a curriculum. 
%\emph{Time to threshold} computes how much faster an agent can learn a policy that achieves return $G_0$ greater than some performance threshold $\delta$ on the target task by following the curriculum, as opposed to learning the target task from scratch. 
%Time can be measured in terms of CPU time, wall clock time, episodes, or number of actions taken; in this work, we will usually use time to refer to number of actions taken. 

An important evaluation question is whether to include time spent \emph{learning in source tasks} into the cost of using transfer. 
%Figure \ref{weakTLImage} shows an example of \emph{weak transfer}. 
The transfer curve in Figure \ref{tlMetrics}a shows performance on the target task, and starts at time 0, even though time has already been spent learning one or more source tasks. Thus, it does not reflect time spent training in source tasks before transferring to the target task. 
This is known in transfer learning as the \emph{weak transfer} setting, where time spent training in source tasks is treated as a sunk cost.  
On the other hand, in the \emph{strong transfer} setting, the learning curves must account for time spent in all source tasks. One way to do this is to offset the curves to reflect time spent in source tasks, as shown in Figure \ref{tlMetrics}b. Another option is to freeze the policy while learning on source tasks, and plot that policy's performance on the target task.

%In weak transfer, time spent training in source tasks is treated as a sunk cost, and not shown in the learning curve graphs. However,  

%In Figure \ref{weakTLImage}, the curve for transfer shows performance versus training time on the target task, after completing training on all sources. However, ... 
%The learning curves start after all source tasks have been learned. I.e. the transfer also starts at 0, but it actually spent many episodes learning in source tasks that is not reflected in the curve. 

%Graphs like these show weak transfer. They don't reflect time spent training in the curriculum or constructing the curriculum. 

\section{The Curriculum Learning Method}
\label{sec:curriculum_problem}
%{\bf TO DO: formulate this in the most general case, i.e., a directed acyclic graph similar to what was in Max's paper - ask Matteo?}

A \emph{curriculum} serves to sort the experience an agent acquires over time, in order to accelerate or improve learning. In the rest of this section we formalize this concept and the methodology of \emph{curriculum learning}, and describe how to evaluate the benefits and costs of using a curriculum. Finally, we provide a list of attributes which we will use to categorize curriculum learning approaches in the rest of this survey. 

\subsection{Curricula}
\label{sec:curricula_def}

A curriculum is a general concept that encompasses both schedules for organizing past experiences, and schedules for acquiring experience by training on tasks. As such, we first propose a fully general definition of curriculum, and then follow it with refinements that apply to special cases common in the literature.

We assume a \emph{task} is modeled as a Markov Decision Process, and define a curriculum as follows:
\begin{definition}[Curriculum]
\label{def:curriculum}
Let $\mathcal{T}$ be a set of tasks, where $m_i = (\mathcal{S}_i, \mathcal{A}_i, p_i, r_i)$ is a task in $\mathcal{T}$. Let $\mathcal{D}^{\mathcal{T}}$ be the set of all possible transition samples from tasks in $\mathcal{T}$: $\mathcal{D}^{\mathcal{T}} = \{ (s, a, r, s') \: | \: \exists \, m_i \in \mathcal{T} \; \mathrm{s.t.} \; s \in \mathcal{S}_{i}, a \in \mathcal{A}_{i}, s' \sim p_i( \cdot | s,  a), r \gets r_i(s, a, s') \}$. 
A \emph{curriculum} $C = (\mathcal{V}, \mathcal{E}, g, \mathcal{T})$ is a directed acyclic graph, where $\mathcal{V}$ is the set of vertices, $\mathcal{E} \subseteq 
\{ (x,y) \; | \; (x,y) \in  \mathcal{V} \times \mathcal{V} \: \land x \neq y \}$ is the set of directed edges, and 
$g : \mathcal{V} \to \mathcal{P}(\mathcal{D}^{\mathcal{T}})$ is a function that associates vertices to subsets of samples in $\mathcal{D}^{\mathcal{T}}$,  where $\mathcal{P}(\mathcal{D}^{\mathcal{T}})$ is the power set of $\mathcal{D}^{\mathcal{T}}$.
%$g : \mathcal{V} \mapsto \mathcal{D}^{\mathcal{T}}_i$ is a function that associates vertices to subsets of samples $\mathcal{D}^{\mathcal{T}}_i \subseteq \mathcal{D}^{\mathcal{T}}$. 
% Vertices $v_i \in \mathcal{V}$ represent subsets of samples from $\mathcal{D}$: $v_i \subseteq \mathcal{D}$. 
A directed edge $\langle v_j, v_k \rangle$ in $C$ indicates that samples associated with $v_j \in \mathcal{V}$ should be trained on before samples associated with $v_k \in \mathcal{V}$. All paths terminate on a single sink node $v_t \in \mathcal{V}$.\footnote{In theory, a curriculum could have multiple sink nodes corresponding to different target tasks. For the purpose of exposition, we assume a separate curriculum is created and used for each task.} 
%a strict partial order over $D$.
\end{definition}

%\comments{Also regarding notation, Sutton also overloads the lowercase $r$. The capital letters are used for RV, which only make sense inside the actual algorithm, not the definition of the VF.}

A curriculum can be created online, where edges are added dynamically based on the learning progress of the agent on the samples at a given vertex. It can also be designed completely offline, where the graph is generated before training, and edges are selected based on properties of the samples associated with different vertices. 

Creating a curriculum graph at the sample level can be computationally difficult for large tasks, or large sets of tasks. Therefore, in practice, a simplified representation for a curriculum is often used. There are 3 common dimensions along which this simplification can happen. The first is the single-task curriculum, where all samples used in the curriculum come from a single task:

%We consider 3 special cases of this definition that are prevalent in existing work. 

%A \emph{curriculum} is an ordered tuple of subsets $\langle \mathcal{D}_1, \mathcal{D}_2, \ldots \mathcal{D}_n\rangle$, where each $\mathcal{D}_i \subseteq \mathcal{D}$ and samples in $\mathcal{D}_i$ are trained on before samples in $\mathcal{D}_j$ for $i < j$. 

% add justification for why partial order for samples covers a wide variety of approaches that order samples individually or in batches

\begin{definition}[Single-task Curriculum]
A \emph{single-task curriculum} is a curriculum $C$ where the cardinality of the set of tasks considered for extracting samples $|\mathcal{T}| = 1$, and consists of only the target task $m_t$. 
\end{definition}

A single-task curriculum essentially considers how best to organize and train on experience acquired from a single task. This type of curriculum is common in experience replay methods \citep{Schaul16}. %, and is similar to how curricula are defined in supervised learning.

%No transfer needed (we haven't introduced this yet though..)
%Defining a curriculum at the sample level may be infeasible for large tasks, or large sets of tasks. Thus, 

A second common simplification is to learn a curriculum   
%In this case, the partial ordering may be defined 
at the task level, where each vertex in the graph is associated with samples from a single task. 
 %which induces a partial order on the samples obtained from the tasks. 
At the task level, a curriculum can be defined as a directed acyclic graph of \emph{intermediate} tasks:

\begin{definition}[Task-level Curriculum]

For each task $m_i \in \mathcal{T}$, let $\mathcal{D}^{\mathcal{T}}_i$ be the set of all samples associated with task $m_i$: $\mathcal{D}^{\mathcal{T}}_i = \{ (s, a, r, s') \: | \: s \in \mathcal{S}_{i}, a \in \mathcal{A}_{i}, s' \sim p_i( \cdot | s,  a), r \gets r_i(s, a, s')  \}$. 
A \emph{task-level curriculum} is a curriculum $C = (\mathcal{V}, \mathcal{E}, g, \mathcal{T})$ where each vertex is associated with samples from a single task in $\mathcal{T}$. Thus, the mapping function $g$ is defined as $g : \mathcal{V} \to \{ \mathcal{D}^{\mathcal{T}}_i  \; | \;  m_i \in \mathcal{T}\}$.

%the mapping function $g : \mathcal{V} \mapsto \mathcal{D}^{\mathcal{T}}_i$ associates each vertex with samples from a single task in $\mathcal{T}$. 
\end{definition}

\begin{figure*}[t]
\begin{center}$
\begin{array}{cc}
\includegraphics[height=5.5cm]{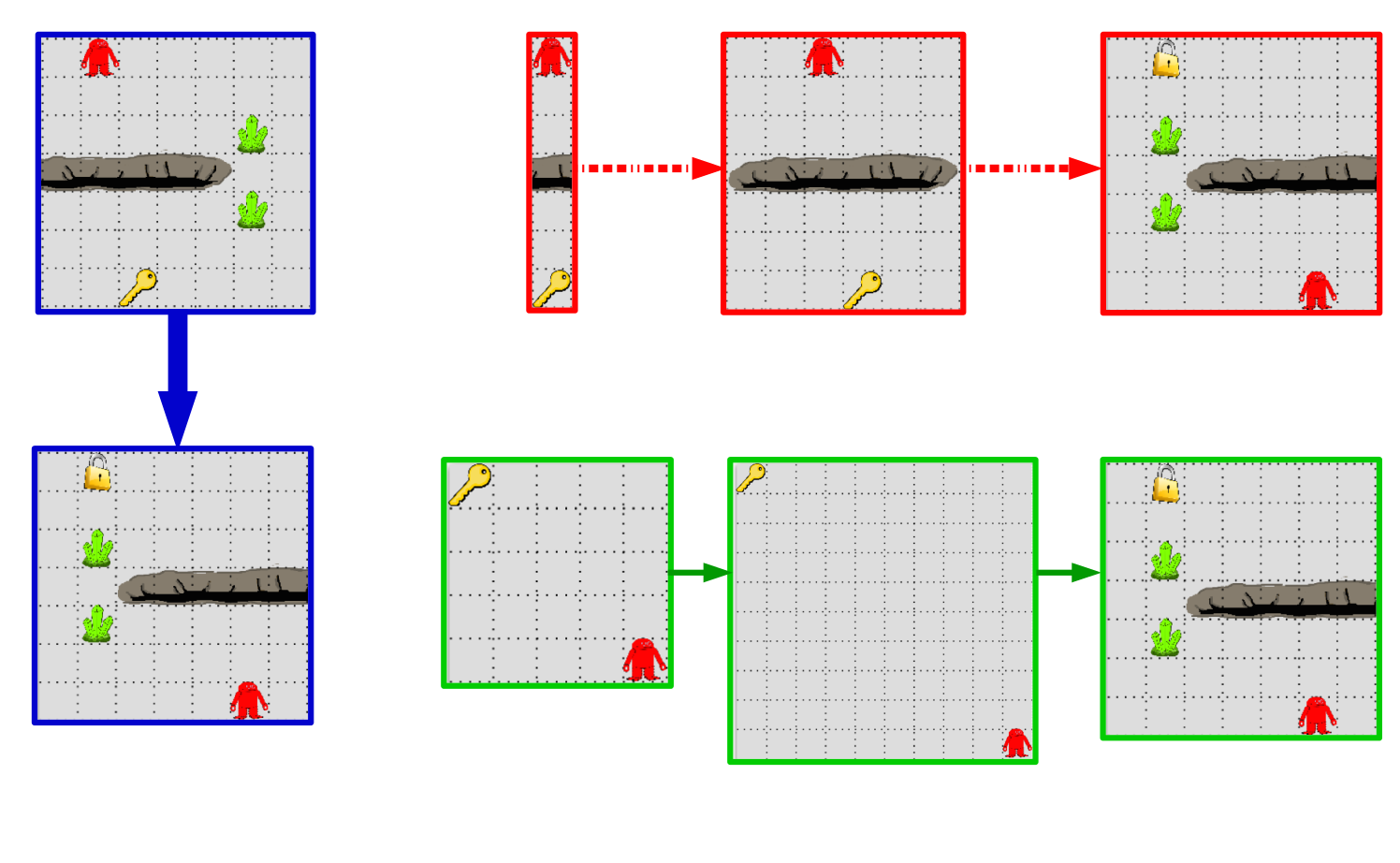} & \includegraphics[height=5.5cm]{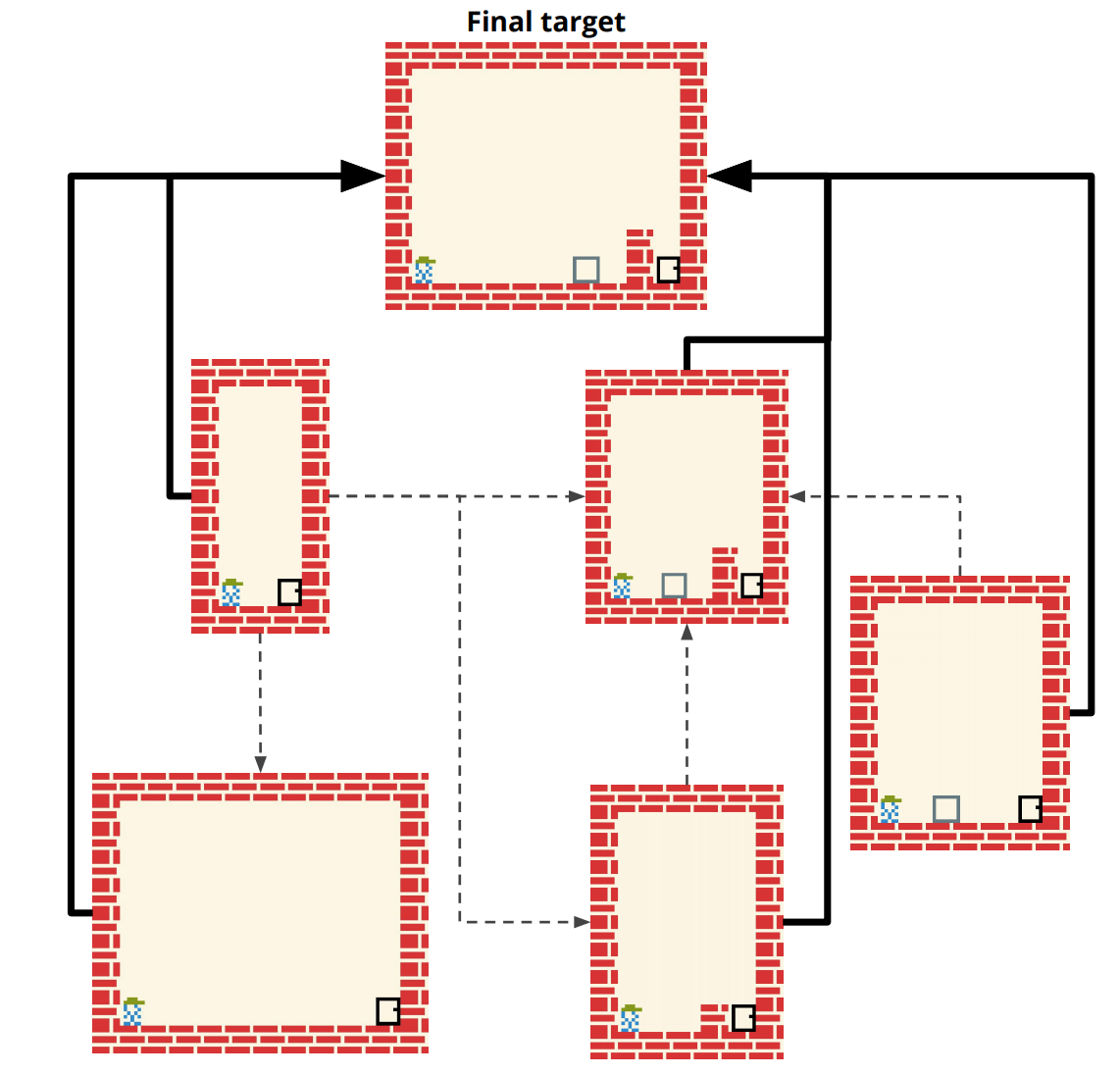} \\
(a) & (b)
\end{array}$
\end{center}
\caption{Examples of structures of curricula from previous work. (a) Linear sequences in a gridworld domain \citep{narvekar2017autonomous} (b) Directed acyclic graphs in block dude \citep{svetlik2017automatic}.}
\label{curriculum_structures}
\end{figure*}

In reinforcement learning, the entire set of samples from a task (or multiple tasks) is usually not available ahead of time. Instead, the samples experienced in a task depend on the agent's behavior policy, which can be influenced by previous tasks learned. Therefore, while generating a task-level curriculum, the main challenge is how to order tasks such that the behavior policy learned is useful for acquiring good samples in future tasks. In other words, selecting and training on a task $m$ induces a mapping function $g$, and determines the set of samples $\mathcal{D}_i^{\mathcal{T}}$ that will be available at the next vertex based on the agent's behavior policy as a result of learning $m$.
The same task is allowed to appear at more than one vertex, similar to how in Definition \ref{def:curriculum} the same set of samples can be associated with more than one vertex. Therefore, tasks can be revisited when the agent's behavior policy has changed. Several works have considered learning task-level curricula over a graph of tasks \citep{svetlik2017automatic,macalpine2018overlapping}. %, which allow transfer from many-to-one and many-to-many tasks. 
An example can be seen in Figure \ref{curriculum_structures}b.

%Typically, the sink node in the graph is the target task. 
%is a directed acyclic graph, where $\mathcal{V}$ is the set of vertices and $E \subseteq \mathcal{V} \times \mathcal{V}$ is the set of edges. Vertices $v \in \mathcal{V}$ represent tasks $m_i \in \mathcal{T}$, with the same task allowed to appear at more than one vertex. An edge $\langle m_i, m_j \rangle$ in $C$ implies that samples from $m_i \in \mathcal{V}$ precede samples from $m_j \in \mathcal{V}$. 

%Thus, one advantage of defining curricula at the task level is that this knowledge of all samples in a task is not required. 
%One advantage of defining a curriculum at the task-level is that it is easier to represent and is more interpretable, because subsets of samples are abstracted away into tasks. 
%Note that the order of the samples within each task is not specified by a task-level curriculum: within tasks, samples can be considered in any order. 

%A key question is how to order tasks, such that samples coll

%Essentially, $g$ serves as a mapping from vertices $\mathcal{V}$ to tasks $\mathcal{T}$.  

Finally, another simplification of the curriculum is the linear \emph{sequence}. This is the simplest and most common structure for a curriculum in existing work:

\begin{definition}[Sequence Curriculum]
A \emph{sequence curriculum} is a curriculum $C$ where the indegree and outdegree of each vertex $v$ in the graph $C$ is at most 1, and there is exactly one source node and one sink node. 
\end{definition}

%\commentj{Why is the sample curriculum a partial order while the task-level one a DAG? Each DAG induces the same partial order but the same order can be represented with possibly more than 1 DAG. Maybe we should explain this right away?}

%In a task-level curriculum $C$, samples from $m_i \in \mathcal{V}$ precede samples from $m_j \in \mathcal{V}$ if there exists an edge $\langle m_i, m_j \rangle$ in $C$. 

These simplifications can be combined to simplify a curriculum along multiple dimensions. For example, the sequence simplification and task-level simplification can be combined to produce a task-level sequence curriculum. This type of curriculum can be represented as an ordered list of tasks $[m_1, m_2, ... m_n]$. An example can be seen in Figure \ref{curriculum_structures}a \citep{narvekar2017autonomous}.

%When vertices are also associated with tasks as in the task curriculum, it is common to 

A final important question when designing curricula is determining the stopping criteria: that is, how to decide \emph{when} to stop training on samples or tasks associated with a vertex, and move on to the next vertex. In practice, typically training is stopped when performance on the task or set of samples has converged. Training to convergence is not always necessary, so another option is to train on each vertex for a fixed number of episodes or epochs. Since more than one vertex can be associated with the same samples/tasks, this experience can be revisited later on in the curriculum.

\subsection{Curriculum Learning}
\label{sec:cl_def}

\emph{Curriculum learning} is a methodology to \emph{optimize} the order in which experience is accumulated by the agent, so as to increase performance or training speed on a set of final tasks. Through generalization, knowledge acquired quickly in simple tasks can be leveraged to reduce the exploration of more complex tasks. 
In the most general case, where the agent can acquire experience from multiple intermediate tasks that differ from the final MDP, there are 3 key elements to this method:
%Two components are therefore central in curriculum learning: experience \emph{sequencing}, and knowledge \emph{transfer}. Furthermore, when curricula are designed at the task-level, a set of intermediate tasks is created for the curriculum, providing a third challenge: \emph{task generation}. 

\begin{itemize}
\item \textbf{Task Generation.} The quality of a curriculum is dependent on the quality of tasks available to choose from. Task generation is the process of creating a good set of intermediate tasks from which to obtain experience samples.  
%When creating a curriculum that utilizes experience from multiple tasks, creating a good set of intermediate tasks is crucial for having positive transfer. 
In a task-level curriculum, these tasks form the nodes of the curriculum graph. This set of intermediate tasks may either be pre-specified, or dynamically generated during the curriculum construction by observing the agent.

\item \textbf{Sequencing.}
Sequencing examines how to create a partial ordering over the set of experience samples $\mathcal{D}$: that is, how to generate the edges of the curriculum graph. Most existing work has used manually defined curricula, where a human selects the ordering of samples or tasks. However, recently automated methods for curriculum sequencing have begun to be explored. Each of these methods make different assumptions about the tasks and transfer methodology used. These methods will be the primary focus of this survey. 

\item \textbf{Transfer Learning.}
When creating a curriculum using multiple tasks, the intermediate tasks may differ in state/action space, reward function, or transition function from the final task. Therefore, transfer learning is needed to extract and pass on reusable knowledge acquired in one task to the next. %While work in transfer learning has typically examined how to transfer knowledge from one or more source tasks directly to the target task, it has generally assumed that both the source and target tasks are prespecified, and that transfer happens in a single stage. 
Typically, work in transfer learning has examined how to transfer knowledge from one or more source tasks directly to the target task. 
Curriculum learning extends the transfer learning scenario to consider training sessions in which the agent must repeatedly transfer knowledge from one task to another, up to a set of final tasks.

\end{itemize}

%, and the set of final tasks may or may not be known at the time of curriculum generation. For instance, developmental approaches \citep{baranes2013active} may both generate tasks online, as the agent progresses throughout the curriculum, and evaluate the learned behavior on a set of tasks unknown until the whole learning process has terminated (an ``adulthood'' phase).

%\emph{Task creation} refers to creating the set of intermediate tasks $\mathcal{T}$ that form the nodes of the curriculum graph. A good set of intermediate tasks is crucial for having positive transfer, and must be relevant to both the domain and the agent in question. 
%Tasks must be \emph{sequenced}, generating the edges of the curriculum graph, in a way that allows new knowledge to be accumulated by the agent along each step.  Finally, between each pair of tasks in the curriculum, \emph{transfer learning}, and appropriate knowledge representations, must be leveraged to transfer and accumulate knowledge.

%The problem of curriculum learning across several tasks relies on knowledge transfer. 
%\comments{Why? Because when you have multiple tasks, they may differ somehow, so you want to transfer what is reusable}

\subsection{Evaluating Curricula}
\label{sec:cl_eval}

Curricula can be evaluated using the same metrics as for transfer learning (cf.~Section \ref{sec:evaluationTL}), by comparing performance on the target task after following the complete curriculum, versus performance following no curriculum (i.e., learning from scratch). %, and averaging the performance over the set of final tasks. 
If there are multiple final tasks, the metrics can easily be extended: for example, by comparing the average asymptotic performance over a set of tasks, or the average time to reach a threshold performance level over a set of tasks. 

Similarly, it is possible to distinguish between weak and strong transfer. However, in curriculum learning, there is the additional expense required to \emph{build} the curriculum itself, in addition to training on intermediate tasks in the curriculum, which can also be factored in when evaluating the cost of the curriculum. 
%Describe how to account for the sunk cost of estimating the curriculum; e.g., strong vs weak transfer.
%In the curriculum learning setting, another question is whether to account for the time spent \emph{constructing} a curriculum. 
As in the transfer learning case, cost can be measured in terms of wall clock time, or data/sample complexity. 

Most existing applications of curricula in reinforcement learning have used curricula created by humans. In these cases, it can be difficult to assess how much time, effort, and prior knowledge was used to design the curriculum.  
Automated approaches to generate a curriculum also typically require some prior knowledge or experience in potential intermediate tasks, in order to guide the sequencing of tasks. Due to these difficulties, these approaches have usually treated curriculum generation as a sunk cost, focusing on evaluating the performance of the curriculum itself, and comparing it versus other curricula, including those designed by people. 
%makes comparisons of the effect of different curricula  and comparing versus other curricula directly. 
%, and potentially compare with curricula created by humans (where it would be harder to assess construction time).

%Describe additional issues to consider: for example, cases where the cost of creating a subtask is expensive relative to the cost of a game step; also, how should human time be values when humans are involved?

The best set of evaluation criteria to use ultimately depends on the specific problem and settings being considered. For example, how expensive is it to collect data on the final task compared to intermediate tasks? If intermediate tasks are relatively inexpensive, we can treat time spent in them as sunk costs. Is it more critical to improve initial performance, final performance, or reaching a desired performance threshold? If designing the curriculum will require human interaction, how will this time be factored into the cost of using a curriculum? Many of these questions depend on whether we wish to evaluate the utility of a specific curriculum (compared to another curriculum), or whether we wish to evaluate the utility of using a curriculum design approach versus training without one. 

\subsection{Dimensions of Categorization}	 
\label{sec:cl_dimensions}

%\comments{TODO} We categorize approaches based on a variety of attributes, discuss them in detail and relate them to the curriculum learning framework described in the previous section. 

We categorize curriculum learning approaches along the following seven dimensions, organized by attributes (in bold) and the values (in italics) they can take. We use these dimensions to create a taxonomy of surveyed work in Section \ref{sec:CLforRLAgents}. %Table \ref{table:categories}	 lists all the articles we survey, and how they fit into our categorization.

%\comments{organized by the attributes and the values they can take}

\begin{enumerate}
\item \textbf{Intermediate task generation}: \emph{target / automatic / domain experts / naive users}. In curriculum learning, the primary challenge is how to sequence a set of tasks to improve learning speed. However, finding a good curriculum depends on first having useful source tasks to select from. Most methods assume the set of possible source tasks is fixed and given ahead of time. In the simplest case, only samples from  the \emph{target} task are used. When more than one intermediate task is used, typically they are manually designed by humans. We distinguish such tasks as designed by either \emph{domain experts}, who have knowledge of the agent and its learning algorithm, or \emph{naive users}, who do not have this information. On the other hand, some works consider \emph{automatically} creating tasks online using a set of rules or generative process. These approaches may still rely on some human input to control/tune hyper-parameters, such as the number of tasks generated, or to verify that generated tasks are actually solvable. %Further reducing the amount of human intervention needed to create source tasks remains an important open problem. 

\item \textbf{Curriculum representation}: \emph{single / sequence / graph}. As we discussed previously, the most general form of a curriculum is a directed acyclic graph over subsets of samples. However, in practice, simplified versions of this representation are often used. 
%there are many ways to represent a curriculum. 
In the simplest case, a curriculum is an ordering over samples from a \emph{single} task. When multiple tasks can be used in a curriculum, curricula are often created at the task-level.  %ordering individual samples from all tasks can be computationally expensive. Thus, one simplifying assumption is to create an ordering over tasks, which induces a partial ordering over the samples. The simplest way to represent a curriculum with multiple tasks this way is 
These curricula can be represented as a linear chain, or \emph{sequence}. In this case, there is exactly one source for each intermediate task in the curriculum. It is up to the transfer learning algorithm to appropriately retain and combine information gathered from previous tasks in the chain. More generally, they can be represented as a full directed acyclic \emph{graph} of tasks. This form supports transfer learning methods that transfer from many-to-one, one-to-many, and many-to-many tasks. %Sequence curricula and graph curricula can be created at the sample-level or the task-level. However, in practice they are usually made at the task-level, and this is what we use for our categorizations. 

%\comments{Is this reasonable as a match to the definitions provided?}

\item \textbf{Transfer method}: \emph{policies / value function / task model / partial policies / shaping reward / other / no transfer}. Curriculum learning leverages ideas from transfer learning to transfer knowledge between tasks in the curriculum. As such, the transfer learning algorithm used affects how the curriculum will be produced. 
The type of knowledge transferred can be low-level knowledge, such as an entire \emph{policy}, an \emph{(action-)value function}, or a full \emph{task model}, which can be used to directly initialize the learner in the target task. It can also be high-level knowledge, such as \emph{partial policies} (e.g.~options) or \emph{shaping rewards}. %, or subtask definitions. 
This type of information may not fully initialize the learner in the target task, but it could be used to guide the agent's learning process in the target task. We use partial policies as an umbrella term to represent closely related ideas such as options, skills, and macro-actions. 
%It is worth noting that options are also referred to as skills and macro-actions. 
When samples from a single task are sequenced, \emph{no transfer} learning algorithm is necessary.
%Most curriculum learning methods are designed for agents that use a particular transfer learning approach. %Some methods are more general, but assume that only one type of transfer learning type is used. 
%However, in the most general case, one would select both the task and what type of information to extract and transfer from it. For example, in some cases, it may be useful to transfer a value function, whereas in others, it may be more beneficial to transfer an option. 
Finally, we use \emph{other} to refer to other types of transfer learning methods. We categorize papers along this dimension based on what is transferred between tasks in the curriculum in each paper's experimental results. %While other transfer methods exist, this list is exhaustive with respect to surveyed work. 

%\comments{Need to give a brief description of what all these mean. And all elements in the table need to appear here. We can probably remove some of this text which appears in open problems, and replace it with a short description of these, and maybe say discussed in more detail in 4.3?}
%\commentn{It seems that people use options, skills, and macro-actions interchangeably. I use partial policies to represent all of them as in Matt's survey paper. Not sure whether you guys think this is the best option.}
%\MET{This is fine with me}

\item \textbf{Curriculum sequencer}: \emph{automatic / domain experts / naive users}. Curriculum learning is a three-part method, consisting of task generation, sequencing, and transfer learning. While much of the attention of this survey is on automated sequencing approaches, many works consider the other parts of this method, and assume the sequencing is done by a human or oracle. %In addition, many papers in different fields such as supervised learning and robotics use a curriculum to aid training. 
Thus, we identify and categorize the type of sequencing approach used in each work similar to task generation: it can be done \emph{automatically} by a sequencing algorithm, or manually by humans that are either \emph{domain experts} or \emph{naive users}.

\item \textbf{Curriculum adaptivity}: \emph{static / adaptive}. Another design question when creating a curriculum is whether it should be generated in its entirety before training, or dynamically adapted during training. We refer to the former type as \emph{static} and to the latter as \emph{adaptive}. Static approaches use properties of the domain and possibly of the learning agent, to generate a curriculum before any task is learned. Adaptive methods, on the other hand, are influenced by properties that can only be measured during learning, such as the learning progress by the agent on the task it is currently facing. For example, learning progress can be used to guide whether subsequent tasks should be easier or harder, as well as how relevant a task is for the agent at a particular point in the curriculum. 

\item \textbf{Evaluation metric}: \emph{time to threshold / asymptotic / jumpstart / total reward}. We discussed four metrics to quantify the effectiveness of learned curricula in Section \ref{sec:cl_eval}. When calculating these metrics, one can choose whether to treat time spent generating the curriculum and training on the curriculum as a sunk cost, or whether to account for both of these for performance. 
Specifically, there are three ways to measure the cost of learning and training via a curriculum. 1) The cost of generating and using the curriculum is treated as a sunk cost, and the designer is only concerned with performance on the target task after learning. This case corresponds to the weak transfer setting. 2) The cost of training on intermediate tasks in the curriculum is accounted for, when comparing to training directly on the target task. This case is most common when it is hard to evaluate the cost of generating the curriculum itself, for example if it was hand-designed by a human. 3) Lastly, the most comprehensive case accounts for the cost of generating the curriculum as well as training via the curriculum. We will refer to the last two as strong transfer, and indicate it by bolding the corresponding metric. Note that achieving asymptotic performance improvements implies strong transfer. 

%\comments{Technically this paragraph also doesn't really describe the values.}

%\item \textbf{Publication status}: \emph{journal / conference / workshop / arXiv}. While we consider rigorously peer reviewed publications that have appeared in journals or top-tier conferences to be the most reliable sources, due to the relative youth of the field, we do also include workshop papers and papers that have only appeared on arXiv.  We indicate the publication status of each paper in the tables throughout the survey.

\item \textbf{Application area}: \emph{toy / sim robotics / real robotics / video games / other}. Curriculum learning methods have been tested in a wide variety of domains. \emph{Toy} domains consist of environments such as grid worlds, cart-pole, and other low dimensional environments. \emph{Sim robotics} environments simulate robotic platforms, such as in MuJoCo. \emph{Real robotics} papers test their method on physical robotic platforms. \emph{Video games} consist of game environments such as Starcraft or the Arcade Learning Environment (Atari). Finally, \emph{other} is used for custom domains that do not fit in these categories. We list these so that readers can better understand the scalability and applicability of different approaches, and use these to inform what methods would be suitable for their own problems.

%Due to the relative infancy of this field, articles that have appeared only on arXiv and are not peer-reviewed are indicated separately.

%In this case, one way to show a benefit of learning via a curriculum is to show that the curriculum works for multiple target tasks or multiple agents, so that the cost of generating the curriculum is amortized. In all of these cases, cost can refer to sample cost or computational complexity. 

%\comments{We need to discuss this -- strong vs weak is orthogonal to asymptotic vs time to threshold.}

%\item \emph{Human study: } We identify which papers involve a human study. \comments{Maybe Bei can say something for this?}
%\comments{The human study papers are in their own section, and I think it might be better served as a subdivision in the table rather than a column.}

\end{enumerate}

%%%%%%%%%%%%%%%%%

\section{Curriculum Learning for Reinforcement Learning Agents}
\label{sec:CLforRLAgents}

%%%%%%%%%%%%%%%%%%

%{\bf TO DO: make a table with columns corresponding to the dimensions we are discussion (e.g., static vs adaptive, structural forms, etc.}

%\commentm{just moved here this sort of outline of the rest, needs a little tying up}

In this section, we systematically survey work on each of the three central elements of curriculum learning: task generation (Section \ref{sec:task_generation}), sequencing (Section \ref{sec:task_sequencing}), and transfer learning (Section \ref{sec:knowledge_transfer}).
%As transfer learning is a well studied field, and task generation 
For each of these subproblems, we provide a table that categorizes work surveyed according to the dimensions outlined in Section \ref{sec:curriculum_problem}. 
The bulk of our attention will be devoted to the subproblem most commonly associated with curriculum learning: sequencing.

\subsection{Task Generation}
\label{sec:task_generation}

Task generation is the problem of creating intermediate tasks specifically to be part of a curriculum. In contrast to the life-long learning scenario, where potentially unrelated tasks are constantly proposed to the agent \citep{thrun98lifelong}, the aim of task generation is to create a set of tasks such that knowledge transfer through them is beneficial. Therefore, all the generated tasks should be relevant to the final task(s) and avoid \emph{negative transfer}, where using a task for transfer hurts performance.  The properties of the research surveyed in this section are reported in Table \ref{table:task_generation}.

Very limited work has been dedicated to formally studying this subproblem in the context of reinforcement learning. All known methods assume the domain can be parameterized using some kind of representation, where different instantiations of these parameters create different tasks. For instance, \citet{AAMAS16-Narvekar} introduce a number of methods to create intermediate tasks for a specific final task. The methods hinge on a definition of a domain as a set of MDPs identified by a \emph{task descriptor}, which is a vector of parameters specifying the \emph{degrees of freedom} in the domain. For example, in the quick chess example (see Section \ref{sec:Intro}), these parameters could be the size of the board, number of pawns, etc. By varying the degrees of freedom and applying task \emph{restrictions}, the methods define different types of tasks. Methods introduced include: \emph{task simplification}, which directly changes the degrees of freedom to reduce the task dimensions; \emph{promising initialization}, which modifies the set of initial states by adding states close to high rewards; \emph{mistake learning}, which rewinds the domain to a state a few steps before a mistake is detected and resumes learning from there; and several other methods. The set of methods determine different kinds of possible tasks, which form a space of tasks in which appropriate intermediate tasks can be chosen.

\citet{silva2018object} propose a similar partially automated task generation procedure in their curriculum learning framework, based on Object-Oriented MDPs. Each task is assumed to have a class \emph{environment} parameterized by a number of attributes. A function, which must be provided by the designer, creates simpler versions of the final task by instantiating the attributes with values that make the tasks easier to solve. 
%Subsequently, objects from other classes are added to the tasks, starting from the object of the smallest class, and drawing the number of objects of the other classes at random. 
For example, continuing the quick chess example, the attributes could be the types of pieces, and the values are the number of each type of piece. 
The presence of different kinds and numbers of objects provide a range of tasks with different levels of difficulty. However, since the generation is mostly random, the designer has to make sure that the tasks are indeed solvable.

\begin{table}
    \makegapedcells
    \centering
    \resizebox{\textwidth}{!}{
\begin{tabular}{|c|c|c|c|c|c|c|c|}
    \hline 
    Citation &  \vtop{\hbox{\strut Intermediate}\hbox{\strut Task}\hbox{\strut Generation}}  & \vtop{\hbox{\strut Curriculum}\hbox{\strut Representation}} & \vtop{\hbox{\strut Transfer}\hbox{\strut Method}}& \vtop{\hbox{\strut Curriculum}\hbox{\strut Sequencer}} & \vtop{\hbox{\strut Curriculum}\hbox{\strut Adaptivity}} & \vtop{\hbox{\strut Evaluation}\hbox{\strut Metric}} & \vtop{\hbox{\strut Application}\hbox{\strut Area}} \\ %\vtop{\hbox{\strut Publication}\hbox{\strut Status}} \\ 
    \hline
    	 
	 ~\cite{silva2018object} & automatic & graph & value function & automatic & static & \textbf{time to threshold}, total reward & toy, video games \\ %conference \\
	 ~\cite{AAMAS16-Narvekar} & automatic & sequence & value function & domain experts & adaptive & \textbf{asymptotic} & video games \\ %conference \\
	 %~\cite{Schmidhuber13} & automatic & sequence & skills & automatic & adaptive & \textbf{asymptotic} & journal \\
	 ~\cite{Schmidhuber13} & automatic & sequence & partial policies & automatic & adaptive & \textbf{asymptotic} & other \\ %journal \\
	 ~\cite{Stone94} & automatic & sequence & other & domain experts & adaptive & time to threshold & other \\ %workshop \\
	  \hline
\end{tabular}
    }
    \caption{The papers discussed in Section \ref{sec:task_generation}, categorized along the dimensions presented in Section \ref{sec:cl_dimensions}. Bolded values under evaluation metric indicate strong transfer.}
\label{table:task_generation}
\end{table}

Generating auxiliary intermediate tasks is a problem that has been studied in non-RL contexts as well. For instance, \citet{Stone94} consider how to semiautomatically create subproblems to aid in learning to solve difficult \emph{planning} problems. Rather than using a static analysis of the domain's properties, they propose to use a partially completed search trajectory of the target task to identify what makes a problem difficult, and suggest auxiliary tasks. For example, if the task took too long and there are multiple goals in the task, try changing the order of the goals. Other methods they propose include reducing the number of goals, creating tasks to solve difficult subgoals, and changing domain operators and objects available for binding.  

Lastly, \citet{Schmidhuber13} introduced Powerplay, a framework that focuses on inventing new problems to train a more and more general problem solver in an unsupervised fashion. The system searches for both a new task and a modification of the current problem solver, such that the modified solver can solve all previous tasks, plus the new one. The search acts on a domain-dependent encoding of the problem and the solver, and has been demonstrated on pattern recognition and control tasks~\citep{Srivastava13}. The generator of the task and new solver is given a limited computational budget, so that it favors the generation of the simplest tasks that could not be solved before. Furthermore, a possible task is to solve all previous tasks, but with a more compact representation of the solver. The resulting iterative process makes the system increasingly more competent at different tasks. The task generation process effectively creates a curriculum, although in this context there are no final tasks, and the system continues to generate pairs of problems and solvers indefinitely, without any specific goal.

\subsection{Sequencing}
\label{sec:task_sequencing}

Given a set of tasks, or samples from them, the goal of sequencing is to order them in a way that facilitates learning. 
%Since curricula can be designed at either the sample or the task level (cf. Section \ref{sec:curriculum_problem}), 
Many different sequencing methods exist, each with their own set of assumptions.
One of the fundamental assumptions of curriculum learning is that we can configure the environment to create different tasks. For the practitioner attempting to use curriculum learning, the amount of control one has to shape the environment affects the type of sequencing methods that could be applicable. 
%In some situations, one may not be able to even use more tasks, while in others, entirely new tasks could be created.
Therefore, we categorize sequencing methods by the degree to which intermediate tasks may differ. %, as well as the transfer learning method used. 
Specifically, they form a spectrum, ranging from methods that simply reorder experience in the final task without modifying any property of the corresponding MDP, to ones that define entirely new intermediate tasks, by progressively adjusting some or all of the properties of the final task.

In this subsection, we discuss the different sequencing approaches. 
First, in Section \ref{sec:sample_sequencing}, we consider methods that reorder samples in the target task to derive a curriculum. Experience replay methods are one such example. 
In Section \ref{sec:colearning}, we examine multi-agent approaches to curriculum generation, where the cooperation or competition between two (typically evolving) agents induces a sequence of progressively challenging tasks, like a curriculum. 
Then, in Section \ref{sec:reward_changes}, we begin describing methods that explicitly use intermediate tasks, starting with ones that vary in limited ways from the target task. In particular, these methods only change the reward function and/or the initial and terminal state distributions to create a curriculum. In Section \ref{sec:no_restrictions}, we discuss methods that relax this assumption, and allow intermediate tasks that can vary in any way from the target task MDP. Finally, in Section \ref{sec:curriculum_with_human}, we discuss work that explores how humans sequence tasks into a curriculum.

%In Section \ref{sec:colearning}, we describe co-training methods, where the agent's experience is influenced by another agent acting either cooperatively or adversarially. Broadly speaking, any methodology in which the experience used to update the agent's parameters is not solely dependent on the agent's own experience falls on this section of the spectrum. %These approaches to curriculum learning are analogous to the original supervised curriculum learning paradigm  \citep{bengio2009curriculum}, which attempts to order the training samples that are used to update a classifier's parameters, in a way that maximizes performance with as little data as possible. 

% Explicit/no-restriction curriculum generation approaches

% Because the MDP is allowed to change, the transfer method is important

\subsubsection{Sample Sequencing} 
\label{sec:sample_sequencing}

% Say something about how this is done in supervised learning
% This section describes the RL analog
% DQNs train by replaying experienced samples from a replay buffer
% The original formulation sampled uniformly. However, more intelligent sampling could result in better performance. 
% Typically accomplished through experience replay

First we consider methods that reorder samples from the final task, but do not explicitly change the domain itself. These ideas are similar to curriculum learning for supervised learning \citep{bengio2009curriculum}, where training examples are presented to a learner in a specific order, rather than completely randomly. \citet{bengio2009curriculum} showed that ordering these examples from simple to complex can improve learning speed and generalization ability. An analogous process can be used for reinforcement learning. For example, many current reinforcement learning methods, such as Deep Q Networks (DQN) \citep{mnih2015human} use a replay buffer to store past state-action-reward experience tuples. At each training step, experience tuples are sampled from the buffer and used to train DQN in minibatches. The original formulation of DQN performed this sampling uniformly randomly. However, as in the supervised setting, samples can be reordered or ``prioritized," according to some measure of usefulness or difficulty, to improve learning. %The measure used distinguishes several methods in this category.  

% The metric for priority or usefullness  

%The metric/properties by which this reordering is done (or )

%\comments{Merge with last paragraph}   In these cases, the dynamics of the domain (i.e., the parameters of the MDP) remain the same throughout the agent's interaction with the environment, and instead, the methods bias the data used to update the agent's parameters, effectively applying an ordering over experience samples.

\begin{table}[t]
    \makegapedcells
    \centering
    \resizebox{\textwidth}{!}{
\begin{tabular}{|c|c|c|c|c|c|c|c|}
    \hline 
    Citation &  \vtop{\hbox{\strut Intermediate}\hbox{\strut Task}\hbox{\strut Generation}}  & \vtop{\hbox{\strut Curriculum}\hbox{\strut Representation}} & \vtop{\hbox{\strut Transfer}\hbox{\strut Method}}& \vtop{\hbox{\strut Curriculum}\hbox{\strut Sequencer}} & \vtop{\hbox{\strut Curriculum}\hbox{\strut Adaptivity}} & \vtop{\hbox{\strut Evaluation}\hbox{\strut Metric}} & \vtop{\hbox{\strut Application}\hbox{\strut Area}}  \\ 
    \hline
    \multicolumn{8}{|l|}{Sample Sequencing (Section \ref{sec:sample_sequencing})}\\
    \hline
     ~\cite{Andrychowicz17} & target & single & no transfer & automatic & adaptive & \textbf{asymptotic}  & sim robotics \\ %conference \\
     ~\cite{Fang19} & target & single & no transfer & automatic & adaptive & \textbf{asymptotic} & sim robotics \\ %conference \\    	 
    	 ~\cite{kim2018screenernet} & target & single & no transfer & automatic & adaptive & \textbf{asymptotic}  & toy, other \\ %arXiv \\
%    	 ~\cite{kumar2020discor} & target & single & no transfer & automatic & adaptive & \textbf{asymptotic} & toy, video games, sim robotics \\ %arXiv \\
    	 ~\cite{lee2019sample} & target & single & no transfer & automatic & adaptive & time to threshold & toy, video games \\ %conference \\  
    	 ~\cite{Ren18} & target & single & no transfer & automatic & adaptive & \textbf{asymptotic}  & video games \\ %journal \\
   	 ~\cite{Schaul16} & target & single & no transfer & automatic & adaptive & \textbf{asymptotic} & video games \\ %conference \\  
    	 \hline
    	 \multicolumn{8}{|l|}{Co-learning (Section \ref{sec:colearning})}\\
    	 \hline
    	 ~\cite{Baker19} & automatic & sequence & policies & automatic & adaptive & \textbf{asymptotic}, time to threshold & other \\ %conference \\
	 ~\cite{Bansal18} & automatic & sequence & policies & automatic & adaptive & \textbf{asymptotic}  & sim robotics \\ %conference \\ 
	 ~\cite{Pinto17} & automatic & sequence & policies & automatic & adaptive & time to threshold  & sim robotics \\ %conference \\
	 ~\cite{Sukhbaatar18} & automatic & sequence & policies & automatic & adaptive & time to threshold, \textbf{asymptotic} & toy, video games \\ %conference \\
	 %~\cite{heess2017emergence} & domain experts & sequence & policies & domain experts & static & time to threshold  & arXiv \\
	 ~\cite{vinyals2019grandmaster} & automatic & sequence & policies & automatic & adaptive & \textbf{asymptotic} & video games \\ %journal \\ 
	 \hline
	 \multicolumn{8}{|l|}{Reward and Initial/Terminal State Distribution Changes (Section \ref{sec:reward_changes})}\\
	 \hline
	 ~\cite{Asada96} & domain experts & sequence & value function & automatic & adaptive & \textbf{asymptotic} & sim/real robotics \\ %journal \\
	 ~\cite{baranes2013active} & automatic & sequence & partial policies & automatic & adaptive & \textbf{asymptotic} & sim/real robotics \\ %journal \\
	 ~\cite{Florensa17} & automatic & sequence & policies & automatic & adaptive & \textbf{asymptotic} & sim robotics \\ %conference \\
	 ~\cite{florensa2018automatic} & automatic & sequence & policies & automatic & adaptive & \textbf{asymptotic} & sim robotics \\ %conference \\
	 ~\cite{ivanovic2019barc} & automatic & sequence & policies & automatic & adaptive & \textbf{asymptotic} & sim robotics \\ %conference \\
	 ~\cite{racaniere2019automated} & automatic & sequence & policies & automatic & adaptive & \textbf{asymptotic} & toy, video games \\ %conference \\
	 ~\cite{Riedmiller18} & domain experts & sequence & policies & automatic & adaptive & time to threshold & sim/real robotics \\ %conference \\
	 %~\cite{baranes2013active} & automatic & sequence & options & automatic & adaptive & \textbf{asymptotic} & journal \\
	 ~\cite{wu2017training} & domain experts & sequence & task model & automatic & both & \textbf{asymptotic} & video games \\ %conference \\
	 \hline
	 \multicolumn{8}{|l|}{No Restrictions (Section \ref{sec:no_restrictions})}\\
	 \hline
	 ~\cite{Bassich20} & domain experts & sequence & policies & automatic & adaptive & \textbf{asymptotic}, \textbf{time to threshold}  & toy \\ %workshop \\
	 ~\cite{silva2018object} & automatic & graph & value function & automatic & static & \textbf{time to threshold}, total reward & toy, video games \\ %conference \\
	  \cite{foglinoICDL19} & domain experts & sequence & value function & automatic & static & time to threshold, \textbf{asymptotic}, total reward  & toy \\ %conference\\
	  \cite{foglinoIJCAI19} & domain experts & sequence & value function & automatic & static & total reward  & toy \\ % conference\\
	  \cite{foglinoWCOpt19} & domain experts & sequence & value function & automatic & static & total reward  & toy \\ %conference\\
	  ~\cite{Jain17} & domain experts & sequence & value function & automatic & adaptive & time to threshold, total reward & toy \\ %workshop \\ 
	 ~\cite{matiisen2017teacher} & domain experts & sequence & policies & automatic & adaptive & \textbf{asymptotic} & toy, video games \\ %journal \\
	  ~\cite{narvekar2017autonomous} & automatic & sequence & value function & automatic & adaptive & \textbf{time to threshold} & toy \\ %conference \\
	  ~\cite{AAMAS19-Narvekar} & domain experts & sequence & value function, shaping reward & automatic & adaptive & \textbf{time to threshold} & toy, video games \\ % conference \\
	  ~\cite{svetlik2017automatic} & domain experts & graph & shaping reward & automatic & static & \textbf{asymptotic}, \textbf{time to threshold} & toy, video games \\ %conference \\
	  \hline
	  \multicolumn{8}{|l|}{Human-in-the-loop Curriculum Generation (Section \ref{sec:curriculum_with_human})}\\
	  \hline
	  %~\cite{stanley:cig05} & domain experts & sequence & skills & domain experts & adaptive & \textbf{asymptotic} & workshop \\  
	 ~\cite{hosu2016playing} & target & single & no transfer & automatic & adaptive & \textbf{asymptotic} & video games \\ %workshop \\
	 ~\cite{khan2011humans} & domain experts & sequence & no transfer & naive users  & static & N/A & other \\ % conference \\
	  ~\cite{macalpine2018overlapping} & domain experts & graph & policies & domain experts & static & \textbf{asymptotic} & sim robotics \\ %journal \\
	 ~\cite{peng2018curriculum} & domain experts & sequence & task model & naive users  & static & \textbf{time to threshold} & other \\ % journal \\
	 ~\cite{stanley:cig05} & domain experts & sequence & partial policies & domain experts & adaptive & \textbf{asymptotic} & video games \\ %workshop \\ 
   	  \hline
\end{tabular}
    }
    \caption{The papers discussed in Section \ref{sec:task_sequencing}, categorized along the dimensions presented in Section \ref{sec:cl_dimensions}. Bolded values under evaluation metric indicate strong transfer.}
\label{table:task_sequencing}
\end{table} 

%\comments{Decide whether prioritized sweeping should be included in here}

The first to do this type of sample sequencing in the context of deep learning were \citet{Schaul16}. They proposed Prioritized Experience Replay (PER), which prioritizes and replays \emph{important} transitions more. Important transitions are those with high expected learning progress, which is measured by their temporal difference (TD) error. Intuitively, replaying samples with larger TD errors allows the network to make stronger updates. 
As transitions are learned, the distribution of important transitions changes, leading to an implicit curriculum over the samples. 
%Two problems with such prioritization are that it can 1) lead to a loss of diversity and 2) could also introduce bias. The loss of diversity is a result of replaying only the samples that have high TD error and ignoring others, leading to overfitting. When the rewards are very stochastic, this overfitting could be even more problematic as the TD targets can be incorrect. 
%\citet{Schaul16} deal with this issue by interpolating between uniform random sampling and greedy prioritization. 
%The second problem is that prioritization also leads to bias, because it changes the distribution of sampling. They propose to deal with this problem by using importance sampling, which is gradually applied, so that initially high-bias updates are annealed over time to lower bias. This annealing through importance sampling of the updates also serves to reduce the step size towards later iterations of the algorithm, aiding in convergence. 

Alternative metrics for priority/importance have been explored as well. \cite{Ren18} propose to sort samples using a complexity index (CI) function, which is a combination of a self-paced prioritized function and a coverage penalty function. The self-paced prioritized function selects samples that would be of appropriate difficulty, while the coverage function penalizes transitions that are replayed frequently. They provide one specific instantiation of these functions, which are used in experiments on the Arcade Learning Environment \citep{bellemare13arcade}, and show that it performs better than PER in many cases. However, these functions must be designed individually for each domain, and designing a broadly applicable domain-independent priority function remains an open problem.  
%\comments{This paper is not that great.}

\citet{kim2018screenernet} consider another extension of prioritized experience replay, where the weight/priority of a sample is jointly learned with the main network via a secondary neural network. The secondary network, called ScreenerNet, learns to predict weights according to the error of the sample by the main network. Unlike PER, this approach is memoryless, which means it can directly predict the significance of a training sample even if that particular example was not seen. Thus, the approach could potentially be used to actively request experience tuples that would provide the most information or utility, creating an online curriculum. 
%\comments{FYI I reviewed this paper for NIPS and it was rejected. It could have some interesting future directions (which I allude to here), but as presented, I (and the other reviewers) didn't see the purpose of learning the attachable neural net.}
%\comments{Matt: For these last 2 papers, maybe mention some shortcomings (it isn't clear why I disliked the papers)}

Instead of using sample importance as a metric for sequencing, an alternative idea is to restructure the training process based on trajectories of samples experienced. For example, 
when learning, typically easy to reach states are encountered first, whereas harder to reach states are encountered later on in the learning cycle. However, in practical settings with sparse rewards, these easy to reach states may not provide a reward signal. 
%one way to do this is to learn easy to reach states first, and learn more difficult to reach states later. The issue with doing this naively from a practical perspective is that in sparse reward settings, easy to reach states may not provide a reward signal. 
Hindsight Experience Replay (HER) \citep{Andrychowicz17} is one method to make the most of these early experiences. HER is a method that learns from ``undesired outcomes," in addition to the desired outcome, by replaying each episode with a goal that was actually achieved rather than the one the agent was trying to achieve. The problem is set up as learning a Universal Value Function Approximator (UVFA) \citep{pmlr-v37-schaul15}, which is a value function $v_\pi(s, g)$ defined over states $s$ and goals $g$ . The agent is given an initial state $s_1$ and a desired goal state $g$. Upon executing its policy, the agent may not reach the goal state $g$, and instead land on some other terminal state $s_T$. While this trajectory does not help to learn to achieve $g$, it does help to learn to achieve $s_T$. Thus, this trajectory is added to the replay buffer with the goal state substituted with $s_T$, and used with an off-policy RL algorithm. 
%In addition, for each transition tuple in the trajectory, a random set of states that appeared after the transition are also added as goal states to the replay buffer. 
HER forms a curriculum by taking advantage of the implicit curriculum present in exploration, where early episodes are likely to terminate on easy to reach states, and more difficult to reach states are found later in the training process.

%Subsequently,  explicitly introduced a curriculum to control the order in which goals are replayed in HER. 

One of the issues with vanilla HER is that all goals in seen trajectories are replayed evenly, but some goals may be more useful at different points of learning. Thus, \cite{Fang19} later proposed Curriculum-guided HER (CHER) to adaptively select goals based on two criteria: curiosity, which leads to the selection of diverse goals, and proximity, which selects goals that are closer to the true goal. Both of these criteria rely on a measure of distance or similarity between goal states. At each minibatch optimization step, the objective selects a subset of goals that maximizes the weighted sum of a diversity and proximity score. They manually impose a curriculum that starts biased towards diverse goals and gradually shifts towards proximity based goals using a weighting factor that is exponentially scaled over time.

Other than PER and HER, there are other works that reorder/resample experiences in a novel way to improve learning. One example is the episodic backward update (EBU) method developed by \cite{lee2019sample}. In order to speed up the propagation of delayed rewards (e.g., a reward might only be obtained at the end of an episode), \cite{lee2019sample} proposed to sample a whole episode from the replay buffer and update the values of all transitions within the sampled episode in a backward fashion. Starting from the end of the sampled episode, the $\max$ Bellman operator is applied recursively to update the target $Q$-values until the start of the sampled episode. This process basically reorders all the transitions within each sampled episode from the last timestep of the episode to the first, leading to an implicit curriculum. Updating highly correlated states in a sequence while using function approximation is known to suffer from cumulative overestimation errors. To overcome this issue, a diffusion factor $\beta \in (0, 1)$ was introduced to update the current $Q$-value using a weighted sum of the new bootstrapped target value and the pre-existing $Q$-value estimate. Their experimental results show that in 49 Atari games, EBU can achieve the same mean and median human normalized performance of DQN by using significantly fewer samples.

Methods that sequence experience samples have wide applicability and found broad success in many applications, since they can be applied directly on the target task without needing to create intermediate tasks that alter the environment. 
%is the simplest version of curriculum learning for RL, since there is only a single task (the final task), and correspondingly no change in the environment. 
In the following sections, we consider sequencing approaches that progressively alter how much intermediate tasks in the curriculum may differ.

%make fewer assumptions on the source and intermediate tasks of the curriculum. 

\subsubsection{Co-learning}
\label{sec:colearning}

%\comments{This section needs more details on how the papers relate to one another. E.g.~what are similarities and differences.}

Co-learning is a multi-agent approach to curriculum learning, in which the curriculum emerges from the interaction of several agents (or multiple versions of the same agent) in the same environment. These agents may act either cooperatively or adversarially to drive the acquisition of new behaviors, leading to an implicit curriculum where both sets of agents improve over time. Self-play is one methodology that fits into this paradigm, and many landmark results such as TD-Gammon \citep{tesauro1995temporal} and more recently AlphaGo \citep{silver2016mastering} and AlphaStar \citep{vinyals2019grandmaster} fall into this category. Rather than describing every work that uses self-play or co-learning, we describe a few papers that focus on how the objectives of the multiple agents can be set up to facilitate co-learning. 

%While settings with both adversarial and cooperative agents are possible in principle, to the best of our knowledge, only settings with adversarial agents have been considered.
 
\citet{Sukhbaatar18} proposed a novel method called asymmetric self-play that allows an agent to learn about the environment without any external reward in an unsupervised manner. 
This method considers two agents, a teacher and a student, using the paradigm of ``the teacher proposing a task, and the student doing it.'' The two agents learn their own policies simultaneously by maximizing interdependent reward functions for goal-based tasks. The teacher's task is to navigate to an environment state that the student will use either as 1) a goal, if the environment is resettable, or 2) as a starting state, if the environment is reversible. In the first case, the student's task is to reach the teacher's final state, while in the second case, the student starts from the teacher's final state with the aim of reverting the environment to its original initial state. The student's goal is to minimize the number of actions it needs to complete the task. The teacher, on the other hand, tries to maximize the difference between the actions taken by the student to execute the task, and the actions spent by the teacher to set up the task. The teacher, therefore, tries to identify a state that strikes a balance between being the simplest goal (in terms of number of teacher actions) for itself to find, and the most difficult goal for the student to achieve. This process is iterated to automatically generate a curriculum of intrinsic exploration.
%The main novelty in this work is in how the teacher defines the goal for the student. 

%The policies are encoded by neural nets, in which the input parameters combine both the current state and the training goal state. After training, the student is evaluated on a target task, which is signaled to the agent by setting the goal to $0$. 
%The experiments show that the teacher's goals act as a shaping reward, and as such they may favor or disadvantage the student on the target task, depending on the adherence of the shaping to the target reward function. If the self-play learns incorrect biases, the agent will have to overcome what has been learned in the training phase. 
%Also for their evaluation, they mix the self-play episodes with target task episodes (that way there is no ``forgetting", etc.), but don't count self-play episodes towards training time cost as it doesn't use the true environment reward.

Another example of jointly training a pair of agents adversarially for policy learning in single-agent RL tasks is Robust Adversarial RL (RARL) by \citet{Pinto17}. Unlike asymmetric self-play \citep{Sukhbaatar18}, in which the teacher defines the goal for the student, RARL trains a protagonist and an adversary, where the protagonist learns to complete the original RL task while being robust to the disturbance forces applied by the adversarial agent. 
RARL is targeted at robotic systems that are required to generalize effectively from simulation, and learn robust policies with respect to variations in physical parameters. Such variations are modeled as disturbances controlled by an adversarial agent, and the adversarial agent's goal is to learn the optimal sequence of destabilizing actions via a zero-sum game training procedure. The adversarial agent tries to identify the hardest conditions under which the protagonist agent may be required to act, increasing the agent's robustness. Learning takes place in turns, with the protagonist learning against a fixed antagonist's policy, and then the antagonist learning against a fixed protagonist's policy. Each agent tries to maximize its own return, and the returns are zero-sum. %, so that if the protagonist gets $r_t$ the adversary gets $-r_t$. 
The set of ``destabilizing actions" available to the antagonist is assumed to be domain knowledge, and given to the adversary ahead of time. 
%I don't see how this process could prevent an adversary from making an impossible task where the protagonist can't learn at all (maybe you have to make the time spent training each round short, and do many rounds). Anyways, you could think of this as an implicit curriculum, because of the turn-by-turn optimization phase where both the protagonist and antagonist are continuously improved.

For multi-agent RL tasks, several works have shown how simple interaction between multiple learning agents in an environment can result in emergent curricula. Such ideas were explored early on in the context of evolutionary algorithms by \cite{rosin1997new}. They showed that competition between 2 groups of agents, dubbed hosts and parasites, could lead to an ``arms race," where each group drives the other to acquire increasingly complex skills and abilities. Similar results have been shown in the context of RL agents by \citet{Baker19}. They demonstrated that increasingly complex behaviors can emerge in a physically grounded task. Specifically, they focus on a game of hide and seek, where there are two teams of agents. One team must hide 
%where 2 teams of agents must either hide from, or seek out agents from the other team, 
with the help of obstacles and other items in the environment, while the other team needs to find the first team. They were able to show that as one team converged on a successful strategy, the other team was pressured to learn a counter-strategy. This process was repeated, inducing a curriculum of increasingly competitive agents. 

A similar idea was explored by \citet{Bansal18}.  
%Further experiments in co-learning which show an emergent curriculum were carried out by \citet{Bansal18}. %They also provide the source code for a number of multi-agent domains. \comments{Maybe we should find somewhere else to discuss source code. It is awkward here.} 
They proposed to use multi-agent curriculum learning as an alternative to engineering dense shaping rewards. Their method interpolates between dense ``exploration'' rewards, and sparse multi-agent competitive rewards, with the exploration reward gradually annealed over time. In order to prevent the adversarial agent from getting too far ahead of the learning agent and making the task impossible, the authors propose to additionally sample older versions of the opponent. Lastly, in order to increase robustness, the stochasticity of the tasks is increased over time.

Curriculum learning approaches have also been proposed for cooperative multi-agent systems \citep{Wang20,Yang20}. In these settings, there is a natural curriculum created by starting with a small number of agents, and gradually increasing them in subsequent tasks. The schedule with which to increase the number of agents is usually manually defined, and the emphasis instead is on how to perform transfer when the number of agents change. Therefore, we discuss these approaches in more detail in Section \ref{sec:knowledge_transfer}.

Finally, while self-play has been successful in a wide variety of domains, including solving games such as Backgammon \citep{tesauro1995temporal} and Go \citep{silver2016mastering}, such an approach alone was not sufficient for producing strong agents in a complex, multi-agent, partially-observable game like Starcraft. One of the primary new elements of \cite{vinyals2019grandmaster} was the introduction of a Starcraft League, a group of agents that have differing strategies learned from a combination of imitation learning from human game data and reinforcement learning. 
% are initialized with strategies learned from imitation learning from human game data. 
Rather than have every agent in the league maximize their own probability of winning against all other agents like in standard self play, there were some agents that did this, and some whose goal was to optimize against the main agent being trained. In effect, these agents were trained to exploit weaknesses in the main agent and help it improve. Training against different sets of agents over time from the league induced a curriculum that allowed the main agents to achieve grandmaster status in the game. 

%\comments{This needs to be cleaned up and added to the table. Also say something about the curriculum here.}

\subsubsection{Reward and Initial/Terminal State Distribution Changes}
\label{sec:reward_changes}

Thus far, the curriculum consisted of ordering experience from the target task or modifying agents in the target environment. In the next two sections, we begin to examine approaches that explicitly create different MDPs for intermediate tasks, by changing some aspect of the MDP. First we consider approaches that keep the state and action spaces the same, as well as the environment dynamics, but allow the reward function and initial/terminal state distributions to vary. 

%\comments{Merge with previous or scrap?} In Section \ref{sec:reward_changes}, we discuss methodologies in which some limited aspects of the domain are changed over the course of training. For example, some approaches modify the reward function so as to guide the agent through tasks that would otherwise have sparse reward functions. Other approaches employ auxiliary tasks -- ones that have the same dynamics as the target task, but different reward functions. Other subtle changes to the MDP include explicit changes in the initial and terminal state distributions.

One of the earliest examples of this type of method was \emph{learning from easy missions}. \citet{Asada96} proposed this method to train a robot to shoot a ball into a goal based on vision inputs. The idea was to create a series of tasks, where the agent's initial state distribution starts close to the goal state, and is progressively moved farther away in subsequent tasks, inducing a curriculum of tasks. %Each subsequent task moves the initial states one ``step" farther away from the goal, forming a curriculum of tasks. 
%first starts close to the goal state, and progressively start the agent farther away. 
%In other words, start the agent one ``step" (or state) away from the goal, then 2 steps, and so on, forming a curriculum of tasks. 
In this work, each new task starts one ``step" farther away from the goal, where steps from the goal is measured using a domain specific heuristic: a state is closer to the terminal state if the goal in the camera image gets larger. 
The heuristic implicitly requires that the state space can be categorized into ``substates," such as goal size or ball position, 
where the ordering of state transitions in a substate to a goal state is known. Thus, each substate has a dimension for making the task simpler or more complex. Source tasks are manually created to vary along these dimensions of difficulty. % \citet{Asada96} propose to complexify along a dimension if the change in the Q-value for that state is below some threshold. 

%For example, in the robot shooting example, one 

Recently, \citet{Florensa17} proposed more general methods for performing this reverse expansion. They proposed reverse curriculum generation, an algorithm that generates a distribution of starting states that get increasingly farther away from the goal. The method assumes at least one goal state is known, which is used as a seed for expansion. Nearby starting states are generated by taking a random walk from existing starting states by selecting actions with some noise perturbation. In order to select the next round of starting states to expand from, they estimate the expected return for each of these states, and select those that produce a return between a manually set minimum and maximum interval. This interval is tuned to expand states where progress is possible, but not too easy. A similar approach by \citet{ivanovic2019barc} considered combining the reverse expansion phase for curriculum generation with physics-based priors to accelerate learning by continuous control agents.

An opposite ``forward" expansion approach has also been considered by \citet{florensa2018automatic}. This method allows an agent to automatically discover different goals in the state space, and thereby guide exploration of the space. They do this discovery with a Generative Adversarial Network (GAN) \citep{goodfellow2014generative}, where the generator network proposes goal regions (parameterized subsets of the state space) and the discriminator evaluates whether the goal region is of appropriate difficulty for the current ability of the agent. Goal regions are specified using an indicator reward function, and policies are conditioned on the goal in addition to the state, like in a universal value function approximator \citep{pmlr-v37-schaul15}. The agent trains on tasks suggested by the generator. In detail, the approach consists of 3 parts: 1) First, goal regions are labelled according to whether they are of appropriate difficulty. Appropriate goals are those that give a return between hyperparameters $R_{min}$ and $R_{max}$. Requiring at least $R_{min}$ ensures there is a signal for learning progress. Requiring less than $R_{max}$ ensures that it is not too easy. 2) They use the labeled goals to train a Goal GAN. 3) Goals are sampled from the GAN as well as a replay buffer, and used for training to update the policy. %In order to save computation on automatic labelling, they use the previous policy's trajectories on the last set of goals as training for the GAN. 
The goals generated by the GAN shift over time to reflect the difficulty of the tasks, and gradually move from states close to the starting state to those farther away.

\citet{racaniere2019automated} also consider an approach to automatically generate a curriculum of goals for the agent, but for more complex goal-conditioned tasks in dynamic environments where the possible goals vary between episodes. 
The idea was to train a ``setter'' model to propose a curriculum of goals for a ``solver'' agent to attempt to achieve. In order to help the setter balance its goal predictions, they proposed three objectives which lead to a combination of three losses to train the setter model: \textit{goal validity} (the goal should be valid or achievable by the current solver), \textit{goal feasibility} (the goal should match the feasibility estimates for the solver with current skill), and \textit{goal coverage} (encourage the setter to choose more diverse goals to encourage exploration in the space of goals). In addition, a ``judge'' model was trained to predict the reward the current solver agent would achieve on a goal (the feasibility of a goal) proposed by the setter. Their experimental results demonstrate the necessity of all three criteria for building useful curricula of goals. They also show that their approach is more stable and effective than the goal GAN method \citep{florensa2018automatic} on complex tasks.

%\comments{Could optionally delete ``The approach consists of 3 parts ..."}

An alternative to modifying the initial or terminal state distribution is to modify the reward function. 
\citet{Riedmiller18} introduce SAC-X (Scheduled Auxiliary Control), an algorithm for scheduling and executing auxiliary tasks that allow the agent to efficiently explore its environment and also make progress towards solving the final task. %\MET{I was confused here - auxiliary tasks are normally additional tasks that are learned in parallel with the "real" task to maximize reward. This makes it sound like these tasks are used temporarily, and then discarded in favor of the "real" task?}
Auxiliary tasks are defined to be tasks where the state, action, and transition function are the same as the original MDP, but where the reward function is different. The rewards they use in auxiliary tasks correspond to changes in raw or high level sensory input, similar to \cite{Jaderberg17}. However, while \cite{Jaderberg17} only used auxiliary tasks for improving learning of the state representation, here they are used to guide exploration, and are sequenced. The approach is a hierarchical RL method: they need to 1) learn intentions, which are policies for the auxiliary tasks, and 2) learn the scheduler, which sequences intention policies and auxiliary tasks. 
To learn the intentions, they learn to maximize the action-value function of each intention from a starting state distribution that comes as a result of following each of the other intention policies. This process makes the policies compatible. %They also assume the auxiliary rewards for all intentions are available to the agent, so that they can all be learned simultaneously off-policy. 
The scheduler can be thought of as a meta-agent that performs sequencing, whose goal is to maximize the return on the target task MDP. The scheduler selects intentions, whose policy is executed on the extrinsic task, and is used to guide exploration. %During this process, all intentions are updated off policy while learning. 
%Once the period ends, a new intention is selected. 

Heuristic-based methods have also been designed to sequence tasks that differ  in their reward functions. One such approach is SAGG-RIAC (Self-Adaptive Goal Generation - Robust Intelligent Adaptive Curiosity) \citep{baranes2013active}. 
%Another approach is by \citet{baranes2013active}, who define the SAGG-RIAC (Self-Adaptive Goal Generation - Robust Intelligent
%Adaptive Curiosity) architecture. %, an architecture based on the previous RIAC architecture \citep{baranes2009riac}, and extend it to select goals in task space, effectively creating new tasks.
%The tasks selected by SAGG-RIAC belong to the same domain (they share the dynamics and the actions of the system) but differ in the goals, i.e.~in the reward function. %The formalism is designed for motion tasks in robotics, and comprises a measure of 
They define \emph{competence} as the distance between the achieved final state and the goal state, and \emph{interest} as the change in competence over time for a set of goals. A region of the task space is deemed more \emph{interesting} than others, if the latest tasks in the region have achieved a high increase in competence. %The task space %(which can also be learned \citep{pere2018unsupervised}) 
%is initially randomly sampled for goals, and then split into regions of maximal interest difference. 
%The approach initially randomly samples the task space for goals, and splits them into regions of maximal interest difference. 
%The process keeps selecting goals 
The approach repeatedly selects goals by first picking a region with a probability proportional to its interest, and then choosing a goal at random within that region. With a smaller probability the system also selects a goal at random over the whole task set or a goal close to a previously unsuccessful task. The bias towards interesting regions causes the goals to be more dense in regions where the competence increases the fastest, creating a curriculum. Because of the stochastic nature of the goal generating process, however, not every task is necessarily beneficial in directly increasing the agent's ability on the target task, but contributes to updating the competence and interest measures. Since the intermediate tasks are generated online as the agent learns, in this approach both sequencing and generation result from the same sampling process.

Finally,~\cite{wu2017training} also consider changing the transition dynamics and the reward functions of the intermediate tasks. They propose a novel framework for training an agent in a partially observable 3D Doom environment. Doom is a First-Person Shooter game, in which the player controls the agent to fight against enemies. %Their framework combines Asynchronous Advantage Actor-Critic (A3C)~\citep{mnih2016asynchronous} with curriculum learning, given that A3C usually takes a long time to find a good solution with very sparse environmental reward. 
In their experiment, they first train the agent on some simple maps with several curricula. Each curriculum consists of a sequence of progressively more complex environments with varying domain parameters (e.g., the movement speed or initial health of the agent). After learning a capable initial task model, the agent is then trained on more complicated maps and more difficult tasks with a different reward function. They also design an adaptive curriculum learning strategy in which a probability distribution over different levels of curriculum is maintained. When the agent performs well on the current distribution, the probability distribution is shifted towards more difficult tasks. 
%\comments{Not sure how well this paper fits in this section}

\subsubsection{No restrictions}
\label{sec:no_restrictions}

% Overview of this section. How do the papers in this section differ from other sections
%What makes this section different from the others?
%Unlike section blah, these approaches blah 
% Talk about formalizations of the problem. Two broad classes: MDPs based, and graph based. Learning the full mdp as well as exact edges on the graphs are intractable/expensive, so each proposes heuristics. 

Next, there is a class of methods that create a curriculum using intermediate tasks, but make no restrictions on the MDPs of these intermediate tasks. We categorize them in three ways by how they address the task sequencing problem: treating sequencing 1) as an MDP/POMDP, 2) as a combinatorial optimization over sequences, and 3) as learning the connections in a directed acyclic task graph. Because there are no limitations on the types of intermediate tasks allowed, some assumptions are usually made about the transfer learning algorithm, and additional information about the intermediate tasks (such as task descriptors) is typically assumed. Finally, we also discuss work on an auxiliary problem to sequencing: how long to spend on each task.

\subsubsection*{MDP-based Sequencing}

The first formalization of the sequencing problem is as a Markov Decision Process. These methods formulate curriculum generation as an interaction between 2 types of MDPs. The first is the standard MDP, which models a \emph{learning agent} (i.e., the student) interacting with a task.  The second is a higher level meta-MDP for the \emph{curriculum agent} (i.e., the teacher), whose goal is to select tasks for the learning agent.

\citet{narvekar2017autonomous} denote the meta-MDP as a curriculum MDP (CMDP), where the state space $\mathcal{S}$ is the set of policies the learning agent can represent. These can be represented parametrically using the weights of the learning agent. %, or as a table of actions for each state in discrete domains. 
The action space $\mathcal{A}$ is the set of tasks the learning agent can train on next. Learning a task updates the learning agent's policy, and therefore leads to a transition in the CMDP via a transition function $p$. Finally, the reward function $r$ is the time in steps or episodes that it took to learn the selected task. Under this model, a curriculum agent typically starts in an initial state corresponding to a random policy for the learning agent. The goal is to reach a terminal state, which is defined as a policy that can achieve some desired performance threshold on the target task, as fast as possible. 

\citet{matiisen2017teacher} consider a similar framework, where the interaction is defined as a POMDP. The state and action spaces of the meta-POMDP are the same as in \citet{narvekar2017autonomous}, but access to the internal parameters of the learning agent is not available. Instead, an observation of the current score of the agent on each intermediate task is given. The reward is the change in the score on the task from this timestep to the previous timestep when the same task was trained on. Thus, while \citet{narvekar2017autonomous} focused on minimizing time to threshold performance on the target task, the design of \cite{matiisen2017teacher} aims to maximize the sum of performance in all tasks encountered. 

While both approaches are formalized as POMDPs, learning on these POMDPs is computationally expensive. Thus, both propose heuristics to guide the selection of tasks. \citet{narvekar2017autonomous} take a sample-based approach, where a small amount of experience samples gathered on the target and intermediate tasks are compared to identify relevant intermediate tasks. The task that causes the greatest change in policy as evaluated on the target task samples is selected. In contrast, \citet{matiisen2017teacher} select tasks where the absolute value of the slope of the learning curve is highest. Thus it selects tasks where the agent is making the most progress or where the agent is forgetting the most about tasks it has already learned. Initially tasks are sampled randomly. As one task starts making progress, it will be sampled more, until the learning curve plateaus. Then another will be selected, and the cycle will repeat until all the tasks have been learned. 
%\comments{Note: \citet{matiisen2017teacher} seems like it is incomplete (hence it is still only on arXiv). Not sure if we want to keep it.}

Subsequently, \cite{AAMAS19-Narvekar} explored whether learning was possible in a curriculum MDP, thus avoiding the need for heuristics in task sequencing. They showed that you can represent a CMDP state using the weights of the knowledge transfer representation. For example, if the agent uses value function transfer, the CMDP state is represented using the weights of the value function. By utilizing function approximation over this state space, they showed it is possible to learn a policy over this MDP, termed a curriculum policy, which maps from the current status of learning progress of the agent, to the task it should learn next. In addition, the approach addresses the question of how long to train on each intermediate task. While most works have trained on intermediate tasks until learning plateaus, this is not always necessary. \cite{AAMAS19-Narvekar} showed that training on each intermediate task for a few episodes, and letting the curriculum policy reselect tasks that require additional time, results in faster learning. However, while learning a curriculum policy is possible, doing so independently for each agent and task is still very computationally expensive. 

\subsubsection*{Combinatorial  Optimization and Search}

% Introduce what combinatorial optimization is and how CL could fit in that method/area

A second way of approaching sequencing is as a combinatorial optimization problem: given a fixed set of tasks, find the permutation that leads to the best curriculum, where best is determined by one of the CL metrics introduced in Section \ref{sec:cl_eval}. Finding the optimal curriculum is a  computationally difficult black-box optimization problem. Thus, typically fast approximate solutions are preferred. %, such as metaheuristic algorithms. 
%\comments{Could we say what complexity class these methods would fall into? NP-hard? NP-complete? Matteo??}
%\comments{According to this http://www2.informatik.uni-osnabrueck.de/knust/class/ scheduling problems can be of all classes, so I wouldn't make a general assertion like that... We can remove the whole sentence if we are uncomfortable with it.}

One such popular class of methods are metaheuristic algorithms, which are heuristic methods that are not tied to specific problem domains, and thus can be used as black boxes. \citet{foglinoICDL19} adapt and evaluate four representative metaheuristic algorithms to the task sequencing problem: beam search \citep{ow1988filtered}, tabu search \citep{glover1998tabu}, genetic algorithms \citep{Goldberg:1989:GAS:534133}, and ant colony optimization \citep{dorigo1991ant}. The first two are trajectory-based, which start at a guess of the solution, and search the neighborhood of the current guess for a better solution. The last two are population-based, which start with a set of candidate solutions, and improve them as a group towards areas of increasing performance. They evaluate these methods for 3 different objectives: time to threshold, maximum return (asymptotic performance), and cumulative return. Results showed that the trajectory-based methods outperformed their population-based counterparts on the domains tested.

While metaheuristic algorithms are broadly applicable, it is also possible to create specific heuristic search methods targeted at particular problems, such as task sequencing with a specific transfer metric objective. \citet{foglinoIJCAI19} introduce one such heuristic search algorithm, designed to optimize for the cumulative return. Their approach begins by computing transferability between all pairs of tasks, using a simulator to estimate the cumulative return attained by using one task as a source for another. The tasks are then sorted according to their potential of being a good source or target, and iteratively chained in curricula of increasing length. The algorithm is anytime, and eventually exhaustively searches the space of all curricula with a predefined maximum length.

\citet{Jain17} propose 4 different online search methods to sequence tasks into a curriculum. Their methods also assume a simulator is available to evaluate learning on different tasks, and use the learning trajectory of the agent on tasks seen so far to select new tasks. The 4 approaches are: 1) Learn each source task for a fixed number of steps, and add the one that gives the most reward. The intuition is that high reward tasks are the easiest to make progress on.  2) Calculate a transferability matrix for all pairs of tasks, and create a curriculum by chaining tasks backwards from the target tasks greedily with respect to it. 3) Extract a feature vector for each task \citep[as in][]{AAMAS16-Narvekar}, and learn a regression model to predict transferability using the feature vector. 4) Extract pair wise feature vectors between pairs of tasks, and learn a regression model to predict transferability.

%In particular, the methods evaluate transferability between tasks using a simulated agent and select the task that provides the best transfer. 
%They have two parts: training on the curriculum source tasks and then training on the target. A budget of $T$ steps is assigned to do both, and each of the tasks are 	``tried" for some $\tau_i < T$ steps. 
%Specifically 
%The first two don't use domain knowledge, whereas the last two use domain knowledge in the form of task descriptors. 

%, and demonstrate the cumulative return scenario on a micro-grid control problem from real-world data. The maximization of cumulative return proved effective both as a method to shape exploration in the final task, and as a way of addressing sim-to-real transfer.

Finally, instead of treating the entire problem as a black box, it has also been treated as a gray box. \citet{foglinoWCOpt19} propose such an approach, formulating the optimization problem as the composition of a white box scheduling problem and black box parameter optimization. The scheduling formulation partially models the effects of a given sequence, assigning a utility to each task, and a penalty to each pair of tasks, which captures the effect on the objective of learning two tasks one after the other. %Effects of longer sequences (beyond pairs) are ignored. 
The white-box scheduling problem is an integer linear program, with a single optimal solution that can be computed efficiently. The quality of the solution, however, depends on the parameters of the model, which are optimized by a black-box optimization algorithm. This external optimization problem searches the optimal parameters of the internal scheduling problem, so that the output of the two chained optimizers is a curriculum that maximizes cumulative return.  %(the objective is equivalently formulated as minimizing regret in the final task). 

\subsubsection*{Graph-based Sequencing}
Another class of approaches explicitly treats the curriculum sequencing problem as connecting nodes with edges into a directed acyclic task graph. Typically, the task-level curriculum formulation is used, where nodes in the graph are associated with tasks. A directed edge from one node to another implies that one task is a source task for another. %Since the number of possible edges is quadratic in the number of tasks,  

Existing work has relied on heuristics and additional domain information to determine how to connect different task nodes in the graph.  
For instance, \citet{svetlik2017automatic} assume the set of tasks is known in advance, and that each task is represented by a task feature descriptor. These features encode properties of the domain. For example, in a domain like Ms.~Pac-Man, features could be the number of ghosts or the type of maze. The approach consists of three parts. First, a binary feature vector is extracted from the feature vector to represent non-zero elements. This binary vector is used to group subsets of tasks that share similar elements. Second, tasks within each group are connected into subgraphs using a novel heuristic called \emph{transfer potential}. Transfer potential is defined for discrete state spaces, and trades off the applicability of a source task against the cost needed to learn it. Applicability is defined as the number of states that a value function learned in the source can be applied to a target task. The cost of a source task is approximated as the size of its state space. Finally, once subgraphs have been created, they are linked together using directed edges from subgraphs that have a set of binary features to subgraphs that have a superset of those features.

 %Transfer potential weighs the utility of a value function learned in a source task versus the cost required to learn it. The utility is measured as the number of states that a value function learned in the source can be applied to a target task. 
%The feature descriptors for different tasks can have different sizes, indicating whether a property of the domain is missing (for example, in Ms.~Pac-Man, a task could have no ghosts). This is exploited to create subgraphs of tasks that share similar properties (e.g.~one subgraph could be created using only tasks that have no ghosts). 
%In order to take advantage of multiple sources for a given target, Svetlik et al.'s (2017) approach relies on reward shaping as a transfer mechanism, where shaping rewards from multiple sources are added together when they have a directed edge to a common next task.

\citet{silva2018object} follow a similar procedure, but formalize the idea of task feature descriptors using an object-oriented approach. The idea is based on representing the domain as an object-oriented MDP, where states consist of a set of objects. A task OO-MDP is specified by the set of specific objects in this task, and the state, action, transition, and reward functions of the task. With this formulation, source tasks can be generated by selecting a smaller set of objects from the target task to create a simpler task. To create the curriculum graph, they adapt the idea of transfer potential to the object-oriented setting: instead of counting the number of states that the source task value function is applicable in, they compare the sets of objects between the source and target tasks. While the sequencing is automated, human input is still required to make sure the tasks created are solvable.

%As in \cite{svetlik2017automatic}, the entire graph is precomputed and then used for training. 

\subsubsection*{Auxiliary Problems}

Finally, we discuss an additional approach that tackles an auxiliary problem to sequencing: how long to spend on each intermediate task in the curriculum. Most existing work trains on intermediate tasks until performance plateaus. However, as we mentioned previously, \citet{AAMAS19-Narvekar} showed that this is unnecessary, and that better results can be obtained by training for a few episodes, and reselecting or changing tasks dynamically as needed.

\citet{Bassich20} consider an alternative method for this problem
%different method for curriculum generation 
based on \emph{progression} functions. Progression functions specify the pace at which the difficulty of the task should change over time. The method relies on the existence of a task-generation function, which maps a desired complexity $c_t \in [0,1]$ to a task of that complexity. The most complex task, for which $c_t = 1$, is the final task. After every episode, the progression function returns the difficulty of the task that the agent should face at that time. The authors define two types of progression functions: fixed progressions, for which the learning pace is predefined before learning takes place; and adaptive progressions, which adjust the learning pace online based on the performance of the agent. Linear and exponential progressions are two examples of fixed progression functions, and increase the difficulty of the task linearly and exponentially, respectively, over a prespecified number of time steps. The authors also introduce an adaptive progression based on a friction model from physics, which increases $c_t$ as the agent's performance is increasing, and slows down the learning pace if performance decreases. Progression functions allow the method to change the task at every episode, solving the problem of deciding how long to spend in each task, while simultaneously creating a continually changing curriculum.

%... In other words, a total ordering / sequence over the tasks is given. 
%The main assumption is that a set of degrees of freedom $F$ is defined for a domain $D$, and can be varied to produce a set of source task MDPs. They also assume 

%describe alternative approaches for performing sequencing.

%\comments{This doesn't really deal with *sequencing* in the sense of how to order, since the ordering is assumed given. Instead, it's more with how long to train on tasks, with a training schedule. Where should we put this?}

\subsubsection{Human-in-the-Loop Curriculum Generation}
\label{sec:curriculum_with_human}
%Subsection on approaches that use humans (both domain experts and everyday users).
%In some cases, a sequence of tasks with increasing difficulty is predefined. In other cases, a dynamic curriculum is generated during the learning process, with increasing difficulty tailored to the current ability of the learner.

Thus far, all the methods discussed in Section \ref{sec:task_sequencing} create a curriculum \emph{automatically} using a sequencing algorithm, which either reorders samples from the final task or progressively alters how much intermediate tasks in the curriculum may differ. %This class of sequencing methods is typically designed by the \emph{algorithmic experts}, who are familiar with the specifics of the learning algorithm. 
%However, relatively little attention has been paid to the ways in which \emph{humans} design curricula. Yet humans have been manually designing curricula for training both biological and artificial agents for thousands of years. 
\citet{bengio2009curriculum} and \citet{taylor2009assisting} both emphasize the importance of better understanding how \emph{humans} approach designing curricula. 
Humans may be able to design good curricula by considering which intermediate tasks are ``too easy'' or ``too hard,'' given the learner's current ability to learn, similar to how humans are taught with the zone of proximal development~\citep{vygotsky78:zpd}. These insights could then be leveraged when designing automated curriculum learning systems. Therefore, in this section, we consider curriculum sequencing approaches that are done \emph{manually} by humans who are either \emph{domain experts}, who have specialized knowledge of the problem domain, or \emph{naive users}, who do not necessarily know about the problem domain and/or machine learning.

One example of having domain experts manually generate the curriculum is the work done by~\cite{stanley:cig05}, in which they explore how to keep video games interesting by allowing agents to change and to improve through interaction with the player. They use the NeuroEvolving Robotic Operatives (NERO) game, in which simulated robots start the game with no skills and have to learn complicated behaviors in order to play the game. The human player takes the role of a trainer and designs a curriculum of training scenarios to train a team of simulated robots for military combat. The player has a natural interface for setting up training exercises and specifying desired goals. An ideal curriculum would consist of exercises with increasing difficulty so that the agent can start with learning basic skills and gradually building on them. In their experiments, the curriculum is designed by several NERO programmers who are familiar with the game domain. %Thus, they have a good understanding about what kind of training scenarios the agent would need to learn progressively sophisticated skills. 
They show that the simulated robots could successfully be trained to learn different sophisticated battle tactics using the curriculum designed by these domain experts. It is unclear whether the human player who is not familiar with the game can design good curriculum.  

A more recent example is by \cite{macalpine2018overlapping}. They use a very extensive manually constructed curriculum to train agents to play simulated robot soccer. The curriculum consists of a training schedule over 19 different learned behaviors. It encompasses skills such as moving to different positions on the field with different speeds and rotation, variable distance kicking, and accessory skills such as getting up when fallen. Optimizing these skills independently can lead to problems at the intersection of these skills. For example, optimizing for speed in a straight walk can lead to instability if the robot needs to turn or kick due to changing environment conditions. Thus, the authors of this work hand-designed a curriculum to train related skills together using an idea called overlapping layered learning. This curriculum is designed using their domain knowledge of the task and agents. 

While domain experts usually generate good curricula to facilitate learning, most existing work does not explicitly explore their curriculum design process.  
It is unclear what kind of design strategies people follow when sequencing tasks into a curriculum. Published research on Interactive Reinforcement Learning~\citep{thomaz2006reinforcement,KCAP09-knox,suay2011effect,knox2012reinforcement,griffith2013policy,subramanian2016exploration,loftin2016learning,2017ICML-Macglashan} has shown that RL agents can successfully speed up learning using human feedback, demonstrating the significant role can humans play in teaching an agent to learn a (near-) optimal policy. This large body of work mainly focuses on understanding how human teachers want to teach the agent and how to incorporate these insights into the standard RL framework. Similarly, the way we define curriculum design strategies still leaves a lot to be defined by human teachers. %These insights could be leveraged when designing automated curriculum learning systems. 
As pointed out by~\cite{bengio2009curriculum}, the notion of simple and complex tasks is often based on human intuition, and there is value in understanding how humans identify ``simple'' tasks. Along these lines, some work has been done to study whether curriculum design is a prominent teaching strategy that naive users choose to teach the agent and how they approach designing curricula.

\begin{figure*}[t]
\begin{center}$
\begin{array}{cc}
\includegraphics[height=2.8cm]{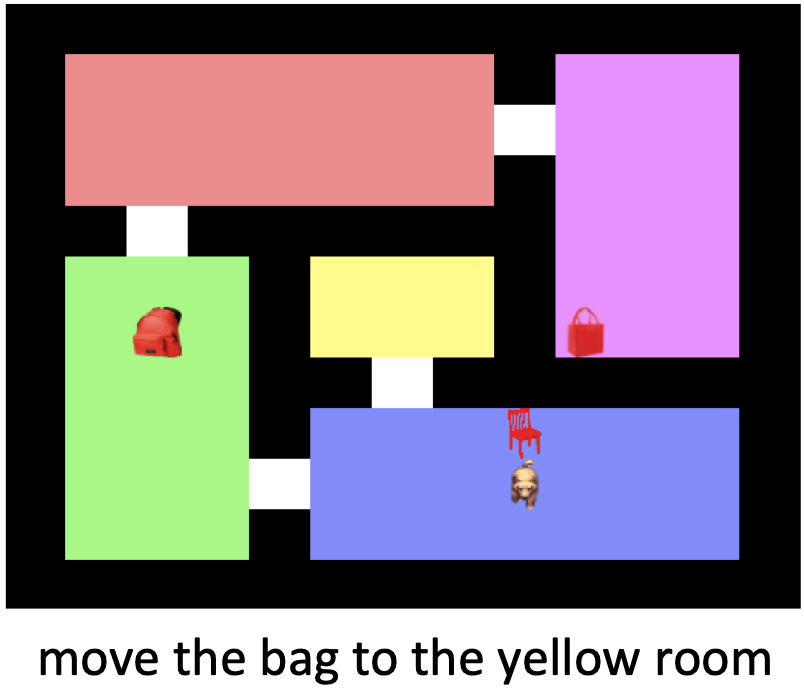} & \includegraphics[width=0.7\columnwidth]{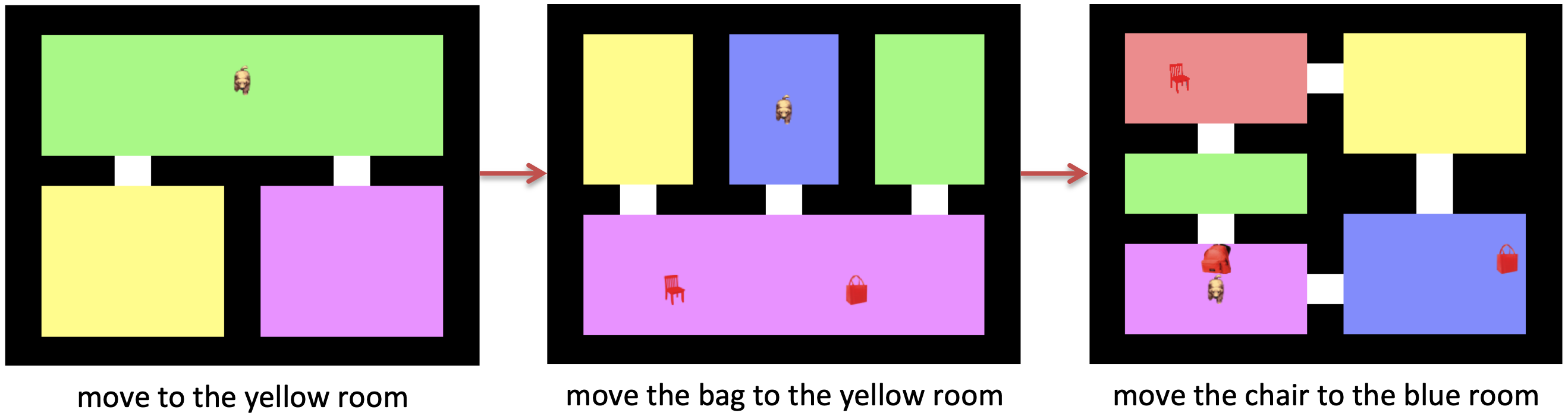} \\
(a) & (b)
\end{array}$
\end{center}
\caption{One example of curricula designed by human users. (a) Given final task. (b) A curriculum designed by one human participant.}
\label{fig:human_curriculum}
\end{figure*}

To study the teaching strategies followed by naive users,~\cite{khan2011humans} conduct behavioral studies in which human participants need to teach a robot the concept of whether an object can be grasped with one hand. In their experiment, participants are provided with 31 cards with photos of common objects (e.g., food, furniture, and animals) for them to select. The experiment consists of two subtasks. In the first subtask, participants sort the objects on the table based on their subjective ratings of their graspability. In the second subtask, participants pick up the cards from the table and show them to the robot while teaching the robot the concept of graspability, using as few cards as possible. While teaching the robot the object's graspability, participants can either use any natural language or say either ``graspable'' or ``not graspable,'' depending on one of the two conditions they are randomly assigned. 
%They observe that participants follow three major teaching strategies: 1) the~\emph{extreme} strategy, which starts with extreme instances that are far away from the decision boundary and gradually moves towards the boundary, 2) the~\emph{linear} strategy, which follows a prominent left-to-right or right-to-left sequence, and 3) the~\emph{positive-only} strategy, which only involves positive examples that participants labeled. The extreme strategy is shown to be consistent with the curriculum learning principle, i.e., starting simple and gradually increasing the difficulty of the task. %\MET{I'm not sure what's meant by "the curriculum learning principle"} 
They observe that participants follow three distinct teaching strategies, one of which is consistent with the curriculum learning principle, i.e., 
starting simple and gradually increasing the difficulty of the task. Furthermore, they propose a novel theoretical framework as a potential explanation for the teaching strategy that follows the curriculum learning principle, which shows that it is the result of minimizing per-iteration expected error of the learner. 

~\cite{peng2018curriculum} also explore how naive users design a curriculum of tasks for an agent, but in a more complex sequential decision-making task. 
%and how to adapt their curriculum-learning algorithms to better take advantage of this type of non-expert guidance. 
Specifically, a simple simulated home environment is used, where the agent must learn to perform tasks in a variety of environments. The tasks are specified via text commands and the agent is trained to perform the task via reinforcement and punishment feedback from a human trainer. It uses the goal-directed Strategy-Aware Bayesian
Learning (SABL) algorithm \citep{loftin2016learning} for learning from human feedback. %Instead of treating the human feedback as a numerical reward signal, SABL interprets it as a discrete communication that depends on both the trainer's desired behavior, and the training strategy they are using. 
In the user study, participants are asked to design a set of training assignments for the agent to help it quickly learn to complete the given final assignment (shown in Figure \ref{fig:human_curriculum}a). A set of source tasks are provided for human participants to select and sequence. %The curriculum is defined to be successful if learning on the given final task is faster with the curriculum than without it. A more difficult goal is to construct a sequence such that training on the entire curriculum and the final task is faster than training directly on the final task. 
One example of curricula designed by human participants is shown in Figure \ref{fig:human_curriculum}b. Their empirical results show that, compared to directly learning the pre-specified final task from scratch, 
non-expert humans can successfully design curricula that result in better overall agent performance on learning both the entire curriculum and the final task. They also discover that humans are more likely to select commands for intermediate tasks that include concepts that are important for the final task, and that doing so results in curricula that lead to better overall agent performance. Furthermore, they demonstrate that by taking advantage of this type of non-expert guidance, %the fact that concepts from the final task are likely to show up in the intermediate tasks,
their curriculum-learning algorithm can be adapted to learn the human-generated curricula more efficiently.
%significantly improved to better handle human-generated curricula. 
 
There is also some work that does not explicitly ask humans to design a curriculum, but uses human data to help generate the curriculum. One example is the work done by~\cite{hosu2016playing}, in which they propose a deep RL method that combines online agent experiences with offline human experiences to train the agent more efficiently. In some sparse-reward Atari games such as Montezuma's Revenge and Private Eye, the agent needs to execute a long sequence of specific actions to receive the first positive reward from the environment, which makes the exploration problem much harder. Thus, the commonly used $\epsilon$-greedy strategy could not find any game paths to reach a first state with positive reward, preventing the neural network from learning relevant features to good states. Inspired by curriculum learning and the human starts evaluation metric used for testing Atari agents, they use checkpoints sampled from a human player's game experience as starting points for the learning process. The main intuition behind this approach is that at least some of the checkpoints will be an ``easier'' starting point, which is closer to some states with positive reward that the agent can benefit from. While this method belongs to the class of sequencing approaches, as discussed in Section \ref{sec:sample_sequencing}, that reorders samples in the final task to derive a curriculum, it additionally considers more informative sample data generated by naive human users in order to build a more efficient curriculum. 

We find that very limited work has been done on investigating how humans design curricula. While the work discussed in this section enriches our empirical understanding of human teaching and gives us some insights into the development of new machine-learning
algorithms and interfaces that can better accommodate machine- or
human-created curricula, we believe more work needs to be done along this line.

\subsection{Knowledge Transfer}
\label{sec:knowledge_transfer}

%\begin{itemize}
%\item no transfer of anything in case of memory ordering, co-learning, and most reward shaping
%\item transfer of reward functions using potential functions
%\item value function transfer
%\item transfer using mappings
%\end{itemize}

While we view sequencing, as covered in Section \ref{sec:task_sequencing}, to be the core concept of curriculum learning, the whole premise of CL depends on an agent's ability to transfer knowledge among tasks.  While a full discussion of transfer learning for RL is beyond the scope of this survey, this subsection is designed to provide the reader a brief introduction to the area so that they can effectively leverage it as part of their own explorations in curriculum learning. 

In curriculum learning, transfer learning methods are used to allow the agent to reuse knowledge learned from one intermediate task to another within the curriculum. It is worth noting that when creating a curriculum using only samples from the target task (discussed in Section~\ref{sec:sample_sequencing}), there is no transfer as there is only a single task (the target task) and correspondingly no change in the environment. %The agent learns from experiences collected through interacting with the same environment. 
However, when creating a curriculum using multiple intermediate tasks, which may differ in state/action space, reward function, or transition function from the final task, transfer learning is needed to extract and pass on reusable knowledge acquired in one intermediate task to the next. The type of knowledge transferred also directly affects the type of learner that is applicable to the learning process. 

Transferred knowledge can be low-level, such as an entire policy, a value function, a full task model, or some training instances, which can be directly used to initialize the learner in the target task. The knowledge can also be high-level, such as partial policies or options, skills, shaping rewards, or subtask definitions. This type of information may not fully initialize the learner in the target task, but it could be used to guide the agent's learning process in the target task. In this subsection, we discuss different transfer learning approaches used in curricula. % and how they deal with multiple tasks in sequence

% Transferred knowledge: policies 
In policy transfer, a policy learned in a source or intermediate task is used to initialize the policy in the target task. When transferring policies between different tasks, the tasks may differ in some aspect of the MDP, such as starting states~\citep{Florensa17}, reward functions~\citep{florensa2018automatic,Riedmiller18}, or transition functions~\citep{clegg2017learning}. For instance,~\cite{clegg2017learning} demonstrate that an arm-like manipulator can successfully learn the control policy for a simulated dressing task, by transferring policies between tasks with different transition functions. In a dressing task, the goal is to achieve a desired relative positioning of the garment and the limb. To do this, they first train a sphere to move through a funnel-like geometry to reach some target location. They then directly apply the learned policy to a different scenario in which a manipulator with arbitrary shape navigates through a simulated garment. The main trick is to train multiple spheres using a curriculum learning strategy and then aggregate them to control the manipulator in the dressing task. 
%Rather than changing the dynamics of the task,~\cite{Florensa17} make the task incrementally harder by moving the agent's start state further and further from the goal state. The agent first learns some locally valid policy for reaching the goal from some start state nearby the goal state. Then, using the policy learned before, the agent backs up to learn what to do in the previous time-step. Here, all start states are automatically generated by executing a short random walk from the previous start states that got some reward but still require more training. \MET{Should we explicitly mention that this is similar to learning from easy missions?}
%\comments{Florensa already appeared before too.. is this critical for a TL section, or should we cut it from here since it's already described?}
%\commentn{Cut this as it is already described in Section 4.2.}

\begin{table}
    \makegapedcells
    \centering
    \resizebox{\textwidth}{!}{
\begin{tabular}{|c|c|c|c|c|c|c|c|}
    \hline 
    Citation &  \vtop{\hbox{\strut Intermediate}\hbox{\strut Task}\hbox{\strut Generation}}  & \vtop{\hbox{\strut Curriculum}\hbox{\strut Representation}} & \vtop{\hbox{\strut Transfer}\hbox{\strut Method}}& \vtop{\hbox{\strut Curriculum}\hbox{\strut Sequencer}} & \vtop{\hbox{\strut Curriculum}\hbox{\strut Adaptivity}} & \vtop{\hbox{\strut Evaluation}\hbox{\strut Metric}} & \vtop{\hbox{\strut Application}\hbox{\strut Area}}  \\ 
    \hline
    	% TODO: think about what papers related to value-function transfer to include in this table
   	  ~\cite{clegg2017learning} & domain experts & sequence & policies & domain experts & static & \textbf{asymptotic}, time to threshold & sim robotics \\ % conference \\  
	  %~\cite{fujii1998multilayered} & domain experts & sequence & options & domain experts & static & \textbf{asymptotic} & conference \\
	  ~\cite{fujii1998multilayered} & domain experts & sequence & partial policies & domain experts & static & \textbf{asymptotic} & real robotics \\ %conference \\
	  %~\cite{vezhnevets2016strategic} & automatic & sequence & macro-actions & automatic & static & \textbf{asymptotic}, total reward & conference \\
	  %~\cite{karpathy2012curriculum} &  domain experts/single   & sequence/sample & skills /no transfer  & domain experts/automatic  &  static/adaptive & time to threshold & conference \\
	  ~\cite{karpathy2012curriculum} &  domain experts/target   & sequence/single & partial policies /no transfer  & domain experts/automatic  &  static/adaptive & time to threshold & sim robotics \\ %conference \\  
	~\cite{Rusu16}  & domain experts & sequence & policies & domain experts & static & \textbf{asymptotic} & video games \\ %arXiv \\
	~\cite{shao2018starcraft} & domain experts & sequence & task model & domain experts & static & \textbf{asymptotic}, total reward & video games \\ %journal \\	
	~\cite{sinapov2015learning}  & automatic & sequence & value function & automatic & static & jump start & video games \\ %conference \\  
	~\cite{tessler2017deep} & domain experts & sequence & partial policies & domain experts & static & \textbf{asymptotic} & video games \\ %conference \\
	~\cite{vezhnevets2016strategic} & automatic & sequence & partial policies & automatic & static & \textbf{asymptotic}, total reward & video games \\ % conference \\
	~\cite{Wang20}  & domain experts & sequence & policies & domain experts & static & \textbf{asymptotic} & video games \\ % conference \\
	 %~\cite{tessler2017deep} & domain experts & sequence & skills & domain experts & static & \textbf{asymptotic} & conference \\
	~\cite{yang1996progressive} & domain experts & sequence & partial policies & automatic & adaptive & \textbf{asymptotic}, \textbf{time to threshold} & real robotics \\ %journal \\
	~\cite{Yang20}  & domain experts & sequence & policies & domain experts & static & \textbf{asymptotic}, time to threshold & toy, other  \\%conference \\
	~\cite{zimmer2018developmental} & domain experts & sequence & partial policies & domain experts & static & \textbf{asymptotic}, total reward & sim robotics \\ %conference \\
   	  \hline
\end{tabular}
    }
    \caption{The papers discussed in Section \ref{sec:knowledge_transfer}, categorized along the dimensions presented in Section \ref{sec:cl_dimensions}. Bolded values under evaluation metric indicate strong transfer.}
\label{table:transfer}
\end{table}

% Transferred knowledge: task models
%\comments{We should probably state what a task model is} \commentn{Not sure whether it's more clear now.}
In~\cite{shao2018starcraft}, a learned task model is transferred between tasks, which is used to initialize the policy network. Thus, it is similar to transferring policies. Their work aims to solve the problem of multi-agent decision making in StarCraft micromanagement, where the goal is to control a group of units to destroy the enemy under certain terrain conditions. 
A parameter sharing multi-agent gradient-descent Sarsa($\lambda $) (PS-MAGDS) method is proposed to train the units to learn an optimal policy, which is parametrized by a feed-forward neural network. PS-MAGDS simply extends the traditional Sarsa($\lambda $) to multiple units by sharing parameters of the policy network among units to encourage cooperative behaviors. A reward function including small immediate rewards is also designed to accelerate the learning process. 
When using transfer learning in their experiments, the agents are first trained in some small scale source scenarios using PS-MAGDS. The well-trained model is then used to initialize the policy network to learn micromanagement in the target scenarios. %\comments{I didn't get this. They learn how to do the target task, and then we transfer in the policy?} \commentn{Fixed.} 
To scale the combat to a large scale scenario, they combine curriculum learning and transfer learning where the agents are trained with a sequence of progressively more complex micromanagement tasks. The difficulty of the micromanagement task is controlled by changing the number and type of units. 

%Similarly,~\cite{wu2017training} gradually adapt the task model learned from pervious tasks to more difficult tasks. In their experiment with First-Person Shooter (FPS) game domain, they first train the agent on some simple map with several curricula. Each curriculum consists of a sequence of progressively more complex environments with varying domain parameters (e.g., the moving speed or initial health of the agent). After learning a capable initial task model, the agent is then trained on more complicated maps and more difficult tasks (e.g., with a different reward function). \commentn{Cut this as it is already described in Section 4.2}

% Transferred knowledge: value function 
Value function transfer is another common method for transferring low-level knowledge between intermediate tasks within a curriculum. In most existing work~\citep{sinapov2015learning, narvekar2017autonomous, silva2018object}, value function transfer is achieved by using the parameters of a value function learned in one intermediate task to initialize the value function in the next intermediate task in the curriculum, such that the agent learns the final task with some initial policy that is better than random exploration. %For instance,~\cite{narvekar2017autonomous} focus on automatically sequencing tasks into a curriculum, assuming access to samples from the final task. They use the learning agent's experience trajectories on the final task to create more relevant intermediate tasks using different heuristics. In contrast,
For example, \cite{sinapov2015learning} focus on addressing the task selection problem in curriculum learning using value function transfer, under the assumption that no samples from the final tasks are available. They propose to use meta-data (i.e., a fixed-length feature vector that describes the task) associated with each task to identify suitable intermediate tasks. The main idea is to use such meta-data to learn the benefits of transfer between different `source-target' task pairs, and have this generalize to new unseen task pairs to guide task selection. %so as to select better source tasks as intermediate tasks. %\comments{Didn't draw the connection between these 2 papers and value function transfer (because the first one talks about samples, and the second says no samples available)...?} \commentn{What I meant here is that both work does value function transfer in the same normal way: use the parameters of a value function learned in the source task to initialize the value function in the target task. But they focus on different transfer problems with different assumptions. I'm not sure whether there is some work that uses value function transfer in a different way.}
%\comments{Brad also commented that the connection between v.f. transfer and the papers mentioned here isn't clear...}

When transferring low-level policies or value functions across tasks, there are several challenges that arise, particularly in the modern context of deep reinforcement learning. First is the problem of catastrophic forgetting, where knowledge from previously learned tasks is lost as information on a new task is incorporated. This effect occurs because the weights of the neural network optimized for a first task must be changed to meet the objectives of a new task, often resulting in poorer performance on the original task. Typically, in the curriculum setting, we only care about performance in the final tasks. However, if information from two orthogonal tasks needs to be combined (such as two independent skills), this challenge needs to be addressed. One approach is progressive neural networks \citep{Rusu16}, which trains a new network ``column" for each new task, and leverages lateral connections to previously learned network columns to achieve transfer. When training subsequent columns, parameters from previous columns are frozen, which prevents catastrophic forgetting. The limitation is that the number of parameters grows with the number of tasks, and at inference time, the task label is needed to know which column to extract output from.

%\comments{In the table, should we apply the curriculum categories, since it is more of a lifelong learning setting?}

% Transferred knowledge: partial policies or options, skills, macro-actions. 
A second problem is the case where the state and action spaces differ between tasks. 
One alternative is to transfer higher-level knowledge across tasks, such as partial policies or options. A partial policy is a policy that is not necessarily defined for all states in the state space of an MDP. We use partial policies as an umbrella term to represent closely related ideas such as options, skills, and macro-actions. ~\cite{yang1996progressive} transfer learned control parameters between tasks, which are similar to partial policies. 
%\comments{Might help to define what partial policies are, and restate how they are similar/different to options and skills?} \commentn{Not sure what is the best definition of partial policy. In this survey, I think we define partial policies, options, and skills as the same thing as mentioned in Section 3.4.}
To solve the impedance learning problem for high-speed robotic assembly, they allow the system to learn impedance parameters associated with different dynamic motions separately, rather than to learn all the control parameters simultaneously. For instance, they first learn only the parameters associated with quasistatic motion by driving the system slowly, leaving other parameters unlearned. After the quasistatic parameters have been learned, they then slightly increase the motion speed, and use the learned values to initialize the quasistatic parameters when learning other parameters. 
Another example of transferring partial policies between tasks is the work done by ~\cite{zimmer2018developmental}. Their main idea is to progressively increase the dimensionality of the tackled problem by increasing the (continuous) state and action spaces of the MDP, while an agent is learning a policy. 
The agent first learns to solve the source task with reduced state and action spaces until the increase in performance stagnates. Then, the partial policy learned by the agent is used as an initialization to learn the full policy in the target task with full state and action spaces. A developmental layer (like a dropout layer) is added to the network to filter dimensions of the states and actions. 

Similarly,~\cite{fujii1998multilayered} transfer options between tasks. To train mobile robots to learn collision avoidance behaviors in multi-robot systems more efficiently, they develop a multi-layered RL mechanism. Rather than gradually increasing the level of task complexity based on the learner's performance as in~\cite{yang1996progressive}, their learning process consists of four stages like a curriculum in which each stage learns a pre-defined controller. Each controller learns an option to solve a pre-defined sub-task. For instance, the first controller learns to move toward a specific goal. Then the output (goal-directed behavior) of the first controller is used as input for the second controller, which aims to learn to avoid the collision to a single robot, and so on. 

\cite{vezhnevets2016strategic} also transfer high-level macro-actions between tasks, which are simpler instances of options. In their experiment, the agent is trained with a curriculum where the goal state is first set to be very close to the start state and is then moved further away during learning process. Although the task gets progressively harder, the temporally abstracted macro-actions remain the same. The macro-actions learned early on can also be easily adapted using their proposed architecture. Specifically, a deep recurrent neural network architecture is used to maintain a multi-step action plan. The network learns when to commit to the action plan to generate macro-actions and when to update the plan based on observations. 

Another mechanism for transfer are skills. ~\cite{tessler2017deep} propose a deep RL method that effectively retains and transfers learned skills to solve lifelong learning in MineCraft. In their work, a set of $N$ skills are trained a priori on various sub-tasks, which are then reused to solve the harder composite task. In their MineCraft experiment, the agent's action space includes the original primitive actions as well as the set of pre-learned skills (e.g., navigate and pickup). A hierarchical architecture is developed to learn a policy that determines when to execute primitive actions and when to reuse pre-learned skills, by extending the vanilla DQN architecture~\citep{mnih2015human}. The skills could be sub-optimal when they are directly reused for more complex tasks, and this hierarchical architecture allows the agent to learn to refine the policy by using primitive actions. They also show the potential for reusing the pre-learned skill to solve related tasks without performing any additional learning. 

Rather than selectively reusing pre-learned skills,~\cite{karpathy2012curriculum} focus on learning motor skills in an order of increasing difficulty. They decompose the acquisition of skills into a two-level curriculum: a~\emph{high-level} curriculum specifies the order in which different motor skills should be learned, while the~\emph{low-level} curriculum defines the learning process for a specific skill. The high-level curriculum orders the skills in a way such that each skill is relatively easy to learn, using the knowledge of the previously learned skills. For instance, the Acrobot first learns the Hop (easy to learn from scratch) and Flip (similar to hopping very slowly) skills, and then learns the more complex Hop-Flip skill. The learned skill-specific task parameters for easier skills will highly constrain the states that the Acrobat could be in, making it easier to learn more complex skills. For example, the Hop-Flip skills begin from a hopping gait of some speed, which can be reached by repeatedly executing the previously learned Hop skill.

In multi-agent settings, several specific methods have been designed for curricula that progressively scale the number of agents between tasks. In these settings, the state and action spaces often scale based on the number of agents present. One common assumption in many of these methods is that the state space can be factored into elements for the environment $s^{env}$, the agent $s^n$, and all other agents $s^{-n}$. For example, \citet{Yang20} propose CM3, which takes a two-stage approach. In the first stage, a single agent is trained without the presence of other agents. This is done by inducing a new MDP that removes all dependencies on agent interactions (i.e.,~removing $s^{-n}$) and training a network on this subspace. Then in the second stage, cooperation is learned by adding the parameters for the other agents into the network. 

\cite{Wang20} propose 3 different approaches for multi-agent settings. The first is buffer reuse, which saves the replay buffers from all previous tasks, and samples experience from all of them to train in the current task. Samples from lower dimensional tasks are padded with zeros. The second is curriculum distillation, which adds a distillation loss based on KL divergence between policies/q-values between tasks. The third is transferring the model using a new network architecture called Dynamic Agent-number Network (DyAN). In this architecture, the state space elements related to the agent and environment go through a fully connected network, while the observations for each teammate agent are passed through a graph neural network (GNN) and then aggregated. These networks are subsequently combined to produce q-values or policies.

\section{Related Areas and Paradigms}
%%%%%%%%%%%%%%%%%%%

\label{sec:related_work}

Curriculum learning is an idea that has been studied in other areas of machine learning and human education, and is similar to several existing paradigms in reinforcement learning. In this section, we first relate curriculum learning to approaches in reinforcement learning that aim to improve sample complexity, and that consider learning multiple sets of tasks (Section \ref{sec:related_paradigms}). Then we describe approaches to learn curricula in supervised learning (Section \ref{sec:cl_for_supervised}) and for teaching and human education (Section \ref{sec:cl_education}). We include these approaches with the idea that the insights discovered in these areas could be adapted to apply to the reinforcement learning setting with autonomous agents. %We also compare how curriculum learning relates to 3 very similar problems well studied in supervised and reinforcement learning. 

%\subsection{Curricula in Developmental Psychology (jivko)}

%Describe how human development from birth to adulthood can be seen as one big curriculum of tasks.

%\begin{itemize}
%\item Piaget and stages of development
%\item Scaffolding in psychology
%\item take home message -- by the time kids start school, their experience has already been highly structured and ordered, i.e., an informal curriculum
%\end{itemize}

%\comments{Are we still keeping this section, or did we decide it is out of scope?}

\subsection{Related Paradigms in Reinforcement Learning}
\label{sec:related_paradigms}

One of the central challenges in applying reinforcement learning to real world problems is sample complexity. Due to issues such as a sparse reward signal or complex dynamics, difficult problems can take an RL agent millions of episodes to learn a good policy, with many suboptimal actions taken during the course of learning. Many different approaches have been proposed to deal with this issue. To name a few, one method is imitation learning \citep{schaal1997learning}, which uses demonstrations from a human as labels for supervised learning to bootstrap the learning process. Another example is off-policy learning \citep{hanna2017data-efficient}, which uses existing data from an observed behavior policy, to estimate the value of a desired target policy. Model-based approaches \citep{Sutton98} first learn a model of the environment, which can then be used for planning the optimal policy. %\comments{Which papers should we cite for these?}

Each of these methods come with their advantages and disadvantages. For imitation learning, the assumption is that human demonstrations are available. However, these are not always easy to obtain, especially when a good policy for the task is not known. In off-policy learning, in order to make full use of existing data, it is assumed that the behavior policy has a nonzero probability of selecting each action, and typically that every action to be evaluated or the target policy has been seen at least once. Finally, model-based approaches typically first learn a model of the environment, and then use it for planning. However, any inaccuracies in the learned model can compound as the planning horizon increases. 
Curriculum learning takes a different approach, and makes a different set of assumptions. The primary assumption is that the environment can be configured to create different subtasks, and that it is easier for the agent to discover \emph{on its own} reusable pieces of knowledge in these subtasks that can be used for solving a more challenging task.

Within reinforcement learning, there are also several paradigms that consider learning on a set of tasks so as to make learning more efficient. Multitask learning, lifelong/continuous learning, active learning, and meta-learning are four such examples. % that are similar to curriculum learning.   %And the automatic construction of curricula can be viewed as a form of active learning. 

In \emph{multitask learning}, the goal is to learn how to solve \emph{sets} of prediction or decision making tasks. Formally, given a set of tasks $m_1, m_2, \ldots m_n$, the goal is to \emph{co-learn} all of these tasks, by optimizing the performance over all $n$ tasks simultaneously. Typically, this optimization is facilitated by learning over some shared basis space. For example, \citet{Caruana97} considers multitask learning for supervised learning problems, and shares layers of a neural network between tasks. In supervised learning, these tasks are different classification or regression problems. Similar ideas have been applied in a reinforcement learning context by \citet{Wilson07}. In reinforcement learning, different tasks correspond to different MDPs.

\emph{Lifelong learning} and \emph{continual learning} can be viewed as an online version of multitask learning. Tasks are presented one at a time to the learner, and the learner must use shared knowledge learned from previous tasks to more efficiently learn the presented task. As in multitask learning, typically the goal is to optimize performance over all tasks given to the learner. Lifelong and continual learning have been examined in both the supervised setting \citep{Ruvolo13} and the reinforcement learning setting \citep{Ring97,Ammar14}. 
%In both of these cases, a set or sequence of tasks is presented to a learner, and the goal is to optimize over all tasks.
The distinguishing feature of curriculum learning compared to these works is that in curriculum learning, we have full control over the \emph{order} in which tasks are 
selected. Indeed, we may have control over the \emph{creation} of tasks as well.  In addition, the goal is to optimize performance for a specific target task, 
rather than all tasks. Thus, source tasks in curriculum learning are designed solely to improve performance on the target task---we are not concerned with optimizing performance in a source.

In \emph{active learning}, the learner chooses which task or example to learn or ask about next, from a given set of tasks. Typically, active learning has been examined in a semi-supervised learning setting: a small amount of labeled data exists whereas a larger amount of unlabeled data is present. The labeled data is used to learn a classifier to infer labels for unlabeled data. Unlabeled data that the classifier is not confident about is requested for a label from a human user. For example, \citet{Ruvolo13b} consider active learning in a lifelong learning setting, and show how a learner can actively select tasks to improve learning speed for all tasks in a set, or for a specific target task. 
The selection of which task to be learned next is similar to the \emph{sequencing} aspect of curriculum learning. However, the full method of curriculum learning is much broader, as it also encompasses creating the space of tasks to consider. \citet{Ruvolo13b} and similar active learning work typically assume the set of tasks to learn and select from are already given. In addition, typically active learning has been examined for supervised prediction tasks, whereas we are concerned with reinforcement learning tasks.

Finally, in \emph{meta-learning} \citep{Finn17}, the goal is to train an agent on a variety of tasks such that it can quickly adapt to a new task within a small number of gradient descent steps. Typically, the agent is not given information identifying the task it is training on. In contrast, in curriculum learning, the learning agent may or may not have information identifying the task. However, the process that designs the curriculum by sequencing tasks usually does have this information. Like in the lifelong setting, there is no significance attached to the order in which tasks are presented to the learner. In addition, the objective in meta-learning is to train for fast adaptability, rather than for a specific final task as is the case in curriculum learning.

% No notion of order/sequencing
% Different objectives -- fast adaptability versus training for a specific target task

\subsection{Curricula in Supervised Machine Learning}
\label{sec:cl_for_supervised}

%Discuss why we are having a small section on this: many of the methods for CL in RL are based on or could benefit from ideas for CL in SL. 
In addition to reinforcement learning, curriculum learning has been examined for supervised learning. While it is beyond the scope of this article to extensively survey supervised CL methods, we would like to highlight a few that could inspire ideas and draw parallels to the RL setting. 
%\comments{Peter says also look up ``stacked generalization" by Wolpert. I don't think this really fits here..}

%There are several differences between curriculum learning for supervised learning versus curriculum learning for reinforcement learning. The main difference between curriculum learning for supervised learning versus reinforcement learning, is that a curriculum is achieved by directly ordering (labeled) training examples. 

\citet{bengio2009curriculum} first formalized the idea of curriculum learning in the context of supervised machine learning.  They conducted case studies examining when and why training with a curriculum can be beneficial for machine learning algorithms, and hypothesized that a curriculum serves as both a continuation method and a regularizer. A continuation method is an optimization method for non-convex criteria, where a smoothed version of the objective is optimized first, with the smoothing gradually reduced over training iterations. Typically, ``easy" examples in a curriculum correspond to a smoother objective. Using a simple shape recognition and language domain, they showed that training with a curriculum can improve both learning speed and performance.

While many papers before \citet{bengio2009curriculum} \emph{used} the idea of a curriculum to improve training of machine learning algorithms, most work considering how to systematically \emph{learn} a curriculum came after. 
One recent example is work by \citet{Graves17}. They introduced measures of \emph{learning progress}, which indicate how well the learner is currently improving from the training examples it is being given. They introduce 2 main measures based on 1) rate of increase in prediction accuracy and 2) rate of increase of network complexity. These serve as the reward to a non-stationary multi-armed bandit algorithm, which learns a stochastic policy for selecting tasks. These signals of learning progress could in theory be applied or adapted to the reinforcement learning setting as well. \citet{Graves17} also make an interesting observation, which is that using a curriculum is similar to changing the step size of the learning algorithm. Specifically, in their experiments, they found that a random curriculum still serves as a strong baseline, because all tasks in the set provide a gradient\footnote{Note however that in the reinforcement learning setting, because the policy affects the distribution of states an agent encounters, random training can be significantly worse.}. Easier tasks provide a stronger gradient while harder tasks provide a gradient closer to 0. Thus, choosing easy, useful tasks allows the algorithm to take larger steps and converge faster.

%L2T paper also poses as RL/MDP. One interesting characterstic about this paper is that they evaluate on a different test distribution. This is harder to do in RL, but would make a stronger case for why to do it. 

More recently, \cite{Fan18} frame curriculum learning as ``Learning to Teach," where a teacher agent learned to train a learning agent using a curriculum. The process is formulated as an MDP between these two interacting agents, similar to the MDP approaches discussed in Section \ref{sec:no_restrictions}: the teacher agent selects the training data, loss function, and hypothesis space, while the learning agent trains given the parameters specified by the teacher. 
The state space of the MDP is represented as a combination of features of the data, features of the student model, and features that represent the combination of both data and learner models. The reward signal is the accuracy on a held-out development set. Training a teacher agent can be computationally expensive. They amortize this cost by using a learned teacher agent to teach a new student with the same architecture. For example, they train the teacher using the first half of MNIST, and use the learned teacher to train a new student from the second half of MNIST. Another way they amortize the cost is to train a new student with a different architecture (e.g., changing from ResNet32 to ResNet110). Similar ideas have been explored in the reinforcement learning setting. However, the test set distribution is different from the training set distribution, which makes performing these kind of evaluations more challenging. However, showing that the cost for training a teacher can be amortized is an important direction for future work.

Finally, \citet{Jiang15} explore the idea of self-paced curriculum learning for supervised learning, which unifies and takes advantage of the benefits of self-paced learning and curriculum learning. In their terminology, curriculum learning uses prior knowledge, but does not adapt to the learner. Specifically, a curriculum is characterized by a ranking function, which orders a dataset of samples by priority. This function is usually derived by predetermined heuristics, and cannot be adjusted by feedback from the learner. 
In contrast, self-paced learning (SPL) adjusts to the learner, but does not incorporate prior knowledge and leads to overfitting. In SPL, the curriculum design is implicitly embedded as a regularization term into the learning objective. However, during learning, the training loss usually dominates over the regularization, leading to overfitting. This paper proposes a framework that unifies these two ideas into a concise optimization problem, and discusses several concrete implementations. The idea is to replace the regularization term in SPL with a self-paced function, such that the weights lie within a predetermined curriculum region. In short, the curriculum region induces \emph{a weak ordering} over the samples, and the self-paced function determines the actual learning scheme within that ordering. The idea has parallels to a task-level curriculum for RL, where the curriculum induces a weak ordering over samples from all tasks, and with the learning algorithm determining the actual scheme within that ordering.

\subsection{Algorithmically Designed Curricula in Education}
\label{sec:cl_education}

% Describe related work in manual and automated curricula generation in education, e.g., intelligent tutoring systems.

% Approaches that use RL to design curricula for people.

%Decades of research in human education have emphasized the role of curriculum design to promote learning. Many researchers in the field focus on designing effective Intelligent Tutoring Systems (ITS) that can teach a student to master multiple skills quickly. 
Curriculum learning has also been widely used for building effective Intelligent Tutoring Systems (ITS) for human education \citep{iglesias2003experience,iglesias2009learning,green2011learning,brunskill2011partially,doroudi2016sequence}.
An ITS system involves a student interacting with an intelligent tutor (a computer-based system), with the goal of helping the student to master all skills quickly, using as little learning content as possible. Given that students have different learning needs, styles, and capabilities, the intelligent tutor should be able to provide customized instructions to them. To achieve this goal, one common strategy is called~\emph{curriculum sequencing}, which aims to provide the learning materials in a meaningful order that maximizes learning of the students with different knowledge levels. The main problem this strategy must solve is to find the most effective lesson to propose next, given the student's current learning needs and capabilities. 

Reinforcement learning is one of the machine learning techniques that has been used with intelligent tutors to partially automate construction of the student model and to automatically compute an optimal teaching policy~\citep{Woolf:2008:BII:2155693}. One advantage of using RL methods in tutoring is that the model can learn adaptive teaching actions based on each individual student's performance in real time, without needing to encode complex pedagogical rules that the system requires to teach effectively (e.g., how to sequence the learning content, when and how to provide an exercise). Another advantage is that it is a general domain-independent technique that can be applied in any ITS.  %it works by simulating a large number of interactions among tutoring polices and simulated students. The policy can be improved through simulations, instead of through real experience, which could be time-consuming and expensive. 

As a concrete example,~\cite{iglesias2003experience, iglesias2009learning} adapt $Q$-learning~\citep{watkins1989learning} to an adaptive and intelligent educational system to allow it to automatically learn how to teach each student. They formulate the learning problem as an RL problem, where the state is defined as the description of the student's knowledge, indicating whether the student has learned each knowledge item. The set of actions the intelligent tutor can execute includes selecting and showing a knowledge item to the student. A positive reward is given when all required content has been learned, otherwise no reward is given. The system evaluates the student's knowledge state through tests, which shows how much the student knows about each knowledge item. The $Q$-value estimates the usefulness of executing an action when the student is in a particular knowledge state. Then, the tutoring problem can be solved using the traditional $Q$-learning algorithm.

%Instead of using a tabular representation of the MDP,
~\cite{green2011learning} propose using a multi-layered Dynamic Bayes Net (DBN) to model the teaching problem in an ITS system. The main idea is to model the dynamics of a student's skill acquisition using a DBN, which is normally used in RL to represent transition functions for state spaces. More specifically, they formulate the problem as a factored MDP, where the state consists of one factor for each skill, corresponding to the student's proficiency on that particular skill. The actions are to either provide a hint or to pose a problem about a particular skill to the student. From a history of teacher-student interaction, the teacher can model the student's proficiency state, with the goal of teaching the student to achieve the highest possible proficiency value on each skill, using as few problems and hints as possible. Subsequently, the learned DBN model is used by a planning algorithm to search for the optimal teaching policy, mapping proficiency states of student knowledge to the most effective problem or hint to pose next. 

To allow the automated teacher to select a sequence of pedagogical actions in cases where learner's knowledge may be unobserved, a different problem formulation is posed by~\cite{rafferty2016faster}. They formulate teaching as a partially observable Markov decision process (POMDP), where the learner's knowledge state is considered as a hidden state, corresponding to the learner's current understanding of the concept being taught. The actions the automated teacher can select is a sequence of pedagogical choices, such as examples or short quizzes. The learner's next knowledge state is dependent on her current knowledge state and the pedagogical action the teacher chooses. Changes in the learner's knowledge state reflect learning. In this framework, the automated teacher makes some assumptions about student learning, which is referred to as the learner model: it specifies the space of possible knowledge states and how the knowledge state changes. Then the teacher can update its beliefs about the learner's current knowledge state based on new observations, given this learner model. Using this POMDP framework, they explore how different learner models affect the teacher's selection of pedagogical actions. 

While most approaches seek to solely maximize overall learning gains,  ~\cite{ramachandran2014adapting} propose an RL-based approach that
uses a personalized social robot to tutor children, that maximizes learning gains and sustained engagement over the student-robot interaction. 
The main goal of the social robot is to learn the ordering of questions presented to a child, based on difficulty level and the child's engagement level in real time. To represent the idea that children with different knowledge levels need a different curriculum, each child is categorized into a given group based on knowledge level at the start of the one-on-one tutoring interaction. An optimal teaching policy is then learned specific to each group. In particular, their approach consists of a training phase and an interaction phase. In the training phase, participants are asked to complete a tutoring exercise. A pretest and post-test will be used to evaluate the participant's relative learning gains, which will also be used as the reward function to learn an optimal policy during the training phase. Subsequently, in the interaction phase, the child's real-time engagement will be detected, serving as another reward signal for the RL algorithm to further optimize the teaching policy. 
%They assume that the system is given $k$ questions with $d$ varying levels of difficulty. It classifies the questions into $d$ different groups based on their difficulty level. Each state represents the difficulty level of a question (question $i$ and $j$ with the same difficulty level $m$ will share the same state $q_{m}$). A transition matrix that contains the state transition probabilities is then constructed. Their approach consists of a training phase and an interaction phase. In the training phase, participants are asked to complete a tutoring exercise, a pretest, and a post-test. The pretest score (representing the user's baseline knowledge) is used to bin each participant into a given category. The difference between the scores of the pretest and post-test serves both as the reward function and learning gains, while the reward function is used to find the optimal policy that provides the ordering of questions presented to the learner. An optimal teaching policy is learned specific to each group, representing the idea that children with different knowledge levels will require a different curriculum. Subsequently, in the interaction phase, the RL model is augmented by customizing the question ordering based on the child's engagement level in real time. In order to detect the child's engagement level, they use a Kinect sensor to characterize the child's confusion based on head tilt and posture changes. \comments{Didn't completely understand this one.}

Non-RL-based algorithms have been considered as well. ~\cite{ballera2014personalizing} leverage the roulette wheel selection algorithm (RWSA) to perform personalized topic sequencing in e-learning systems. RWSA is typically used in genetic algorithms to arrange the chromosomes based on their fitness function, such that individuals with higher fitness value will have higher probability of being selected~\citep{Goldberg:1989:GAS:534133}. Similarly, in an e-learning system, a chromosome is denoted by a lesson. Each lesson has a fitness value that dynamically changes based on the student's learning performance. This fitness value indicates how well the topic was learned by the student, depending on three performance parameters: exam performance, study performance, and review performance of the learner. A lower fitness value means that the student has a poorer understanding of the topic. Thus, a reversed mechanism of RWSA is implemented, so as to select the lessons with lower fitness values more often for reinforcement. Then, this reversed RWSA algorithm is combined with linear ranking algorithm to sort the lessons.

\section{Open Questions}
\label{sec:open_problems}

Through our survey of the literature, we have identified several open problems that have not been sufficiently studied in past work, and could be useful avenues for future research.

%\comments{Maybe make each open problem a subsection, and put a short paragraph or 2 describing the problem, why it's interesting, what existing work has done towards this problem if any, how they fall short, and maybe some ideas, etc?}

\subsection{Fully Automated Task Creation}

Task creation is an important piece of the method of curriculum learning. Whether tasks are created ``on-demand" or all in advance, the quality of the pool of tasks generated directly affects the quality of curricula that can be produced. In addition, the \emph{quantity} of tasks produced affect the search space and efficiency of curriculum sequencing algorithms. Despite this, very limited work (see~Section \ref{sec:task_generation}) has been done on the  problem of automatically generating tasks. Existing work either assumes the pool of tasks are manually crafted and specified beforehand, or defines a set of rules for semi-automatically creating tasks. However, these rules often have hyper-parameters that control how many tasks are created, and are also usually manually tuned. Reducing the amount of manual input required by these methods remains an important area for future work.

\subsection{Transferring Different Types of Knowledge}

Between each pair of tasks in a curriculum, knowledge must be transferred from  one task to the subsequent task. In virtually all of the works surveyed, the type of knowledge transferred has been fixed. For example, a value function was always transferred between tasks by \cite{narvekar2017autonomous} while a shaping reward was always transferred by \cite{svetlik2017automatic}. However, this limitation opens the question of whether different tasks could benefit from extracting different types of knowledge. For instance, it may be useful to extract an option from one task, and a model from another. Thus, in addition to deciding \emph{which} task to transfer from, we could also ask \emph{what} to extract and transfer from that task. Past transfer learning literature has shown that many forms of transfer are possible. %, including value functions \citep{Taylor05}, options \citep{Soni06}, policies \citep{Fernandez10}, models \citep{Fachantidis13}, and samples \citep{Lazaric08, Lazaric11}. 
The best type of knowledge to extract may differ based on task, and techniques will need to be developed to effectively combine these different types of knowledge.

\subsection{Reusing Curricula and Sim-to-Real Curriculum Learning}

Another limitation of many curriculum learning approaches is that the time to generate a curriculum can be greater than the time to learn the target task outright. This shortcoming stems from the fact that curricula are typically learned independently for each agent and target task. However, in areas such as human education, curricula are used to train multiple students in multiple subjects. Thus, one way to amortize the cost would be to learn a curriculum to train multiple different agents, or to solve multiple different target tasks \citep{narvekar2020generalizing}. % Some very preliminary work on generalizing curricula to different tasks has been done by .

%Prior work has shown that agents with different state or action representations can benefit from a customized curriculum. Therefore, a curriculum learned for one agent or task would likely have to be adapted in some way for a new agent or task.

Another option for amortizing the cost is to learn curricula for a sim-to-real setting on physical robots, where a curriculum is learned in simulation and then used to train a physical robot. While the exact weights of the policy learned in simulation would not apply in the real world, the semantics of the curriculum tasks might. Therefore, the physical robot could go through the same training regimen, but learn using the physics and dynamics of the real world.

\subsection{Combining Task Generation and Sequencing}

The curriculum learning method can be thought of as consisting of 3 parts: task generation, sequencing, and transfer learning. For the most part, previous work has tackled each of these pieces independently. For example, sequencing methods typically assume the tasks are prespecified, or a task generation method exists. However, an interesting question is whether the task generation and task sequencing phases can be done simultaneously, by directly generating the next task in the curriculum. Some very preliminary work has been done in this direction in the context of video game level generation. For example, \cite{Green19} used an evolutionary algorithm to generate maps for a gridworld, where each tile had a different element. The generator was optimized to maximize the loss of deep RL agent's network, inducing a training curriculum. 

Combining task generation and sequencing has additional challenges, such as specifying the space of possible maps, ensuring those maps are valid/solvable, and creating maps that are challenging, but not too difficult to solve. In addition, training the generator can be very expensive. However, it promises an end-to-end solution that could reduce the amount of human intervention needed to design curricula.

\subsection{Theoretical Results}

There have been many practical applications of curricula to speed up learning in both supervised and reinforcement learning. However, despite empirical evidence that curricula are beneficial, there is a lack of theoretical results analyzing when and why they are useful, and how they should be created. An initial analysis in the context of supervised learning was done by \cite{Weinshall18} and \cite{Weinshall18b}. They analyzed whether reordering samples in linear regression and binary classification problems could improve the ability to learn new concepts. They did this analysis by formalizing the idea of an Ideal Difficulty Score (IDS), which is the loss of the example with respect to the optimal hypothesis, and the Local Difficulty Score (LDS), which is the loss of the example with respect to the current hypothesis. These are 2 ways to classify the difficulty of a sample, which can be used as a means to sequence samples. They showed that the convergence of an algorithm like stochastic gradient descent monotonically decreases with the IDS, and monotonically increases with the LDS. An open question is whether similar grounded metrics for difficulty of tasks can be identified in reinforcement learning, and what kind of convergence guarantees we can draw from them.

\subsection{Understanding General Principles for Curriculum Design}

%\comments{Bei maybe you can fill this one out? This was in the list of bullets on this section.} \commentn{Done.}

Determining the difficulty of a training example for an agent, and ensuring that each example presented to the agent is suitable given its current ability, is a major challenge in curriculum learning. In most existing work, the curriculum is generated either automatically (see Section \ref{sec:task_sequencing}), by ordering samples from the target tasks or iteratively selecting intermediate tasks with increasing difficulty tailored to the current ability of the learner; or manually by domain experts, who will typically have specialized knowledge of the problem domain. Very limited work (see Section \ref{sec:curriculum_with_human}) has been done to better understand how non-expert humans design curricula. The way we define curriculum design strategies still leaves a lot to be defined by human teachers. 

Can non-expert humans design effective curricula for a given final task? What kind of curriculum design strategies do they tend to follow when building curricula? If we could find some general principles non-expert humans follow for designing and/or sequencing more ``interesting'' intermediate tasks into a curriculum, we could incorporate these insights into the automatic process of generating useful source tasks for any task domain. %The interface design could also be improved to guide the non-experts to design better curricula. 
Furthermore, can we adapt curriculum learning algorithms to better take advantage of this type of non-expert guidance to learn more efficiently? We believe a better understanding of the curriculum-design strategies used by non-expert humans may help us to 1) understand the general principles that make some curriculum strategies work better than others, and 2) inspire the design of new machine-learning algorithms and interfaces that better accommodate the natural tendencies of human trainers.

\section{Conclusion}

This survey formalized the concept of a curriculum, and the method of curriculum learning in the context of reinforcement learning. Curriculum learning is a 3-part approach consisting of 1) task generation, 2) sequencing, and 3) transfer learning. We systematically surveyed existing work addressing each of these parts, with a particular focus on sequencing methods. We broke down sequencing methods into five categories, based on the assumptions they make about intermediate tasks in the curriculum. The simplest of these are sample sequencing methods, which reorder samples from the final task itself, but do not explicitly change the domain. These were followed by co-learning methods, where a curriculum emerges from the interaction of several agents in the same environment. 
Next we considered methods that explicitly changed the MDP to produce intermediate tasks. Some of these assumed that the environment dynamics stay the same, but that the initial/terminal state distribution and reward function can change. Others made no restrictions on the differences allowed from the target task MDP. Finally, we also discussed how humans approach sequencing, to shed light on manually designed curricula in existing work.  
Our survey of the literature concluded with a list of open problems, which we think will serve as worthwhile directions for future work. As a budding area in reinforcement learning, we hope that this survey will provide a common foundation and terminology to promote discussion and advancement in this field.

\acks{We would like to sincerely thank Brad Knox, Garrett Warnell, and the anonymous reviewers for helpful comments and suggestions that improved the presentation of many ideas in this article. 
Part of this work has taken place in the Learning Agents Research
Group (LARG) at the Artificial Intelligence Laboratory, The University
of Texas at Austin. LARG research is supported in part by grants from
the National Science Foundation (CPS-1739964, IIS-1724157,
NRI-1925082), the Office of Naval Research (N00014-18-2243), Future of
Life Institute (RFP2-000), Army Research Office (W911NF-19-2-0333),
DARPA, Lockheed Martin, General Motors, and Bosch.  The views and 
conclusions contained in this document are those of the authors alone.
Peter Stone serves as the Executive Director of Sony AI America and 
receives financial compensation for this work.  The terms of this
arrangement have been reviewed and approved by the University of Texas
at Austin in accordance with its policy on objectivity in research.
Part of this work has taken place in the Sensible Robots Research Group at the University of Leeds, which is partially supported
by the Engineering and Physical Sciences Research Council of the UK (EP/R031193/1, EP/S005056/1), and the British Council.
Part of this work has taken place in the Control, Robotics, Identification and Signal Processing (CRISP) Laboratory at Tufts University which is partially supported by DARPA (W911NF-19-2-0006), the Verizon Foundation, PTC Inc., and the Center for Applied Brain and Cognitive Sciences (CABCS).
Part of this work has taken place in the Whiteson Research Lab at the University of Oxford, which is partially supported by the European Research Council (ERC), under the European Union's Horizon 2020 research and innovation programme (grant agreement number 637713).
Part of this work has taken place in the Intelligent Robot Learning (IRL) Lab at the University of Alberta, which is supported in part by research grants from the Alberta Machine Intelligence Institute.}
\newpage

%\appendix
%\section*{Appendix A.}
%\label{}

% Note: in this sample, the section number is hard-coded in. Following
% proper LaTeX conventions, it should properly be coded as a reference:

%In this appendix we prove the following theorem from
%Section~\ref{sec:textree-generalization}:

\end{document}